\documentclass[10pt,twocolumn,letterpaper]{article}

\usepackage[pagenumbers]{cvpr} 

\definecolor{cblue}{rgb}{0.21,0.49,0.74}
\usepackage[pagebackref,breaklinks,colorlinks,allcolors=cblue]{hyperref}

\title{
UniGeoSeg: Towards Unified Open-World Segmentation for Geospatial Scenes
}

\author{
Shuo Ni$^{1,3}$, Di Wang$^{2,3}$, He Chen$^{1}$, Haonan Guo$^{2,3\dagger}$, Ning Zhang$^{1,4\dagger}$, Jing Zhang$^{2\dagger}$ \\ [2pt]
$^{1}$ Beijing Institute of Technology,
$^{2}$ Wuhan University,\\
$^{3}$ Zhongguancun Academy,
$^{4}$ Hong Kong Polytechnic University \\
\small\textcolor{cblue}{
\texttt{
\href{mailto:shuoni@bit.edu.cn}{shuoni@bit.edu.cn};
\href{mailto:haonan.guo@whu.edu.cn}{haonan.guo@whu.edu.cn};
\href{mailto:nzhang.rs@bit.edu.cn}{nzhang.rs@bit.edu.cn}; 
\href{mailto:jingzhang.cv@gmail.com}{jingzhang.cv@gmail.com}
}}}

\begin{document}
\maketitle
\renewcommand\thefootnote{$\dagger$}
\footnotetext{Corresponding authors.}%
\begin{abstract}
Instruction-driven segmentation in remote sensing generates masks from guidance, offering great potential for accessible and generalizable applications. 
However, existing methods suffer from fragmented task formulations and limited instruction data, hindering effective understanding and generalization.
To address these issues, we introduce \textbf{GeoSeg-1M}, the first million-scale dataset for remote sensing instruction-driven segmentation, constructed via an automatic mask filtering and instruction generation pipeline that synthesizes referring, interactive, and reasoning segmentation instructions from multiple public datasets. GeoSeg-1M contains 590K images, 117 categories, and 1.1M image–mask–instruction triplets.
Building upon this foundation, we further curate \textbf{GeoSeg-Bench}, a challenging benchmark designed to evaluate contextual understanding and reasoning capabilities across diverse instruction-driven tasks and complex geospatial scenes.
Furthermore, we present \textbf{UniGeoSeg}, a unified framework that serves as a strong baseline, incorporating task-aware text enhancement, latent knowledge memory, and a progressive training strategy to facilitate multi-task learning. 
Extensive experiments demonstrate the state-of-the-art performance of UniGeoSeg across GeoSeg-Bench and diverse public benchmarks, while exhibiting strong zero-shot generalization. 
The datasets and source code will be publicly released in \url{https://github.com/MiliLab/UniGeoSeg}.
\end{abstract}    

\section{Introduction}
\label{sec:intro}

\begin{figure}[htbp]
    \centering
    \includegraphics[width=0.9\linewidth]{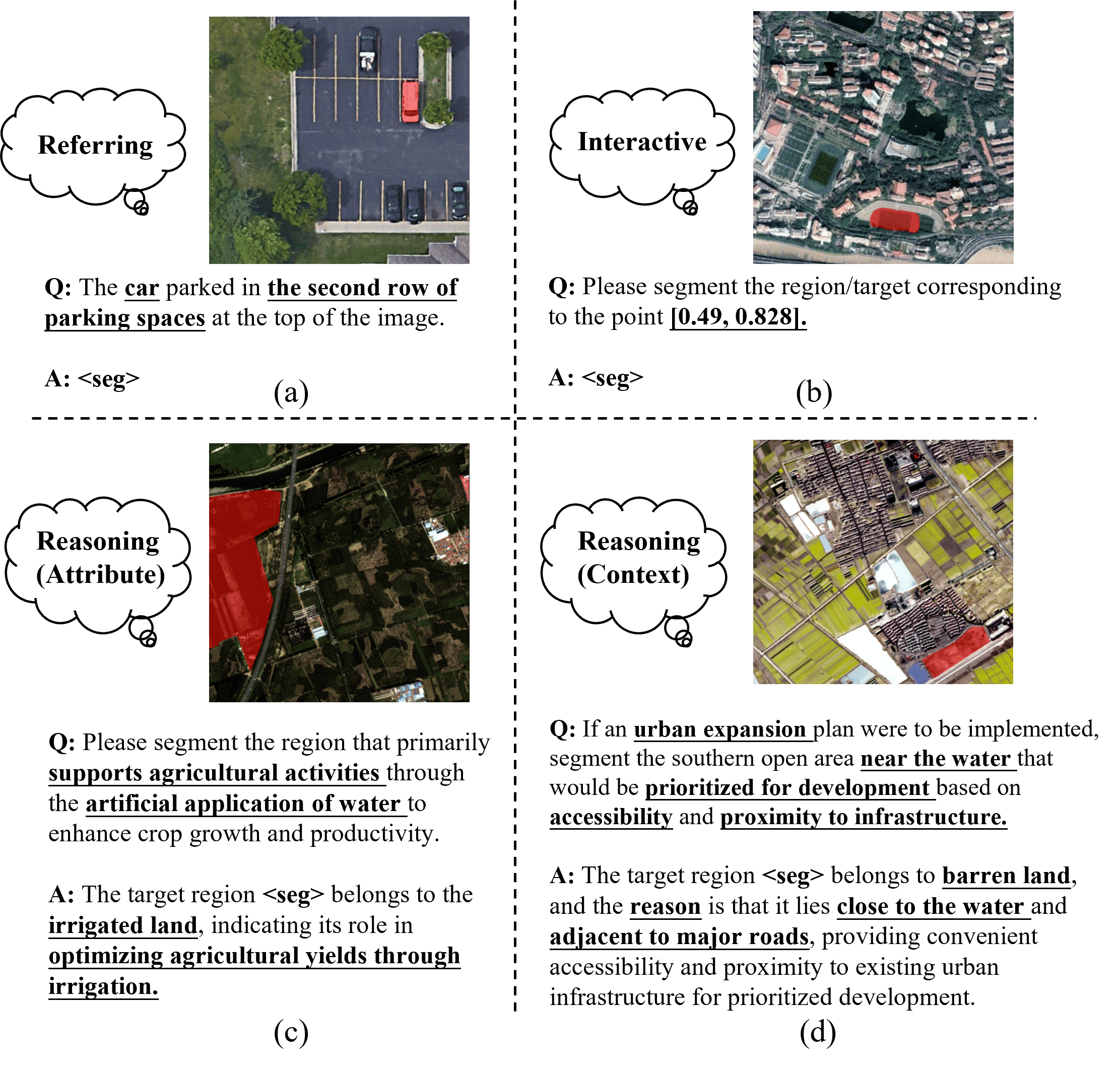}
    \caption{Examples from GeoSeg-1M. (a) Referring segmentation; (b) Interactive segmentation; (c) Attribute-oriented reasoning segmentation; (d) Context-oriented reasoning segmentation.}
    \label{fig:example}
\end{figure}

Instruction-driven segmentation aims to generate pixel-level masks guided by natural language instructions, enabling intuitive and human-centric interaction with visual data. In the field of remote sensing (RS), this paradigm greatly improves the accessibility and practicality of geospatial analysis by enabling flexible segmentation of instruction-specified regions.
It holds broad potential for applications such as urban planning \cite{gao2024enrich,wang2025deep}, environmental monitoring \cite{savva2025advances,wangdgsolver, TiMo}, and disaster assessment \cite{wang2025xlrs,ahn2025generalizable, hypersigma}. 

Recent studies have explored instruction-driven segmentation for RS. For instance, RRSIS-D \cite{liu2024rotated} achieves referring segmentation via multimodal alignment, while SegEarth-R1 \cite{li2025segearthr1} introduces geographic reasoning segmentation. Inspired by the Segment Anything Model (SAM) \cite{kirillov2023segment}, several methods incorporate visual prompting to enhance flexibility and enable adaptive guidance \cite{wang2023samrs,zhang2024earthmarker}. 
Despite these efforts, research on instruction-driven segmentation remains constrained by fragmented task formulations, hindering comprehensive instruction understanding. 
Specifically, existing models often focus on single-purpose segmentation without leveraging the complementary among different tasks, thereby limiting cross-task transferability \cite{wang2025residual,chen2025rsrefseg,lu2025rrsecs, mtp}. Moreover, current datasets lack sufficient data scale and diversity in both visual and textual domains, impeding robust generalization to complex geospatial instructions \cite{ou2025geopix,shabbir2025geopixel, crossearth}. Consequently, these models struggle with open-world understanding, particularly when reasoning over intricate contextual information \cite{pan2025locate,yao2025remotereasoner, s5}.

To address these challenges, we introduce \textbf{GeoSeg-1M}, a large-scale, multi-task instruction-driven segmentation dataset for holistic and comprehensive geospatial understanding. The dataset comprises 590K RS images, 117 semantic categories, and 1.1M image–mask–instruction triplets, covering a wide spectrum of real-world scenes and object types. Some examples are shown in Fig.~\ref{fig:example}.
To construct GeoSeg-1M, we design a comprehensive and automated pipeline for mask filtering and instruction generation that automatically produces referring, interactive, and reasoning segmentation instructions in a unified manner. By leveraging multiple large language models (LLMs) and integrating heterogeneous public datasets, this pipeline constructs context-rich and semantically aligned supervision, effectively bridging language, vision, and spatial reasoning to facilitate multi-task learning.
Furthermore, to provide a consistent evaluation protocol and better assess model understanding and reasoning abilities in complex geospatial environments, we curate GeoSeg-Bench, a high-quality, challenging benchmark designed to assess instruction-driven segmentation in real scene across diverse tasks including referring, interactive, and reasoning.

Building upon this foundation, we further propose \textbf{UniGeoSeg}, a unified framework that serves as a strong baseline for RS instruction-driven segmentation. 
To seamlessly adapt to diverse segmentation scenarios and bridge different task types, UniGeoSeg incorporates three key components: a task-aware text enhancement module that refines instruction comprehension across heterogeneous tasks, a latent knowledge memory module that facilitates semantic sharing among tasks, and a progressive training strategy that incrementally increases task difficulty to enhance generalization.
Extensive experiments verify that UniGeoSeg achieves state-of-the-art results on GeoSeg-Bench and other public benchmarks, and exhibits strong zero-shot adaptability and open-world generalization.
Our main contributions are summarized as follows:
\begin{itemize}

\item We introduce GeoSeg-1M, the first million-scale instruction-driven segmentation RS dataset. It unifies referring, interactive, and reasoning segmentation tasks with 1.1M image–mask–instruction triplets. Besides, we curate a challenging benchmark GeoSeg-Bench emphasizing spatial and contextual reasoning.

\item We propose UniGeoSeg, a unified framework for instruction-driven segmentation, which integrates task-aware text enhancement, latent knowledge memory module, and progressive training to facilitate comprehensive multi-task representation and understanding.

\item Extensive and comprehensive experiments demonstrate that UniGeoSeg achieves state-of-the-art performance across GeoSeg-Bench and multiple public benchmarks, serving as a strong foundation for instruction-driven geospatial segmentation research.

\end{itemize}

\section{Related Work}
\label{sec:formatting}
\subsection{Instruction-Driven Segmentation}

Early methods \cite{hu2016segmentation,li2018referring,shi2018key} typically extracted visual and textual features separately with convolutional and recurrent networks, followed by feature fusion for segmentation. Transformer-based approaches \cite{yang2022lavt,wu2024toward,nag2024safari} further improved cross-modal semantic alignment via attention mechanisms. SAM \cite{kirillov2023segment} introduced visual prompting, enabling open-world and interactive segmentation. LISA \cite{lai2024lisa} incorporated large multimodal models and proposed reasoning-based segmentation tasks, while PSALM \cite{zhang2024psalm} designed well-structured input schemas to support multiple tasks, including video referring segmentation and open-vocabulary segmentation. However, due to the substantial semantic gap between natural images and RS imagery \cite{lu2025rrsecs}, these methods often fail to generalize well on geospatial scenes.

\subsection{Instruction-Driven Segmentation for RS}
To enable instruction-driven in the geospatial domain, Yuan et al.~\cite{yuan2024rrsis} constructed the first referring segmentation dataset tailored to RS imagery. Liu et al. \cite{liu2024rotated} subsequently built RRSIS-D based on RSVG dataset \cite{zhan2023rsvg} and SAM. SegEarth-OV \cite{li2025segearth} leveraged a CLIP-based \cite{radford2021CLIP}, training-free architecture to achieve open-vocabulary segmentation. GeoPixel \cite{shabbir2025geopixel} introduced grounded conversation generation by incorporating powerful LLMs, enabling instruction-guided segmentation with richer textual context. 
More recently, Segearth-R1 \cite{li2025segearthr1} further proposed geospatial pixel reasoning, requiring world knowledge to complete segmentation tasks. Despite these advances, existing instruction-driven segmentation methods remain largely task-specific, limiting the capability to handle complex instructions across diverse geospatial scenarios.

\begin{table*}[thbp]
\centering
\small
\caption{Comparison of GeoSeg-1M with existing RS multimodal segmentation datasets.}
\label{tab:dataset_comparison}
\resizebox{\textwidth}{!}{%
\begin{tabular}{lccccccccc}
\toprule
Dataset & Images & Samples & Avg Text Length & Categories & Spatial Resolution & Dataset Resource$^1$  & Task Types \\
\midrule
RefSegRS\cite{yuan2024rrsis}  & 285 & 4.4 K & 3.09 words &14  & 0.5 m -- 30 m & S & Referring \\
RRSIS-D\cite{liu2024rotated}  & 17 K & 17 K & 6.8 words &20 & 0.13 m & R, DI  & Referring  \\
RISBench\cite{dong2024cross}  & 29 K & 52 K & 15.3 words &26 & 0.1 m -- 30 m & DO, DI  & Referring \\
RemoteSAM\cite{yao2025remotesam}  & 71 K & 270 K & 6.9 words & 297 & 0.05 m -- 30 m & \makecell[c]{ DO, DI, Vi, L, P, S, H} & Referring \\
EarthReason\cite{li2025segearthr1}  & 5 K & 14 K & 20.89 words & 28 & 0.5 m -- 153 m & A, FM  & Reasoning \\
\midrule
GeoSeg-1M & 590 K & 1,149 K &\makecell[c]{12.18 words\\ (12.05/9.80/23.93)$^2$} &117 & 0.05 m -- 153 m &
\makecell[c]{C, DG, FB, FL, GI,\\ GL, L, M, P, Va, FA \\ DO, DI, EeathReason,\\RemoteSAM, RRSIS-D} &
\makecell[c]{Referring/\\ Interactive/\\ Reasoning} \\
\bottomrule
\end{tabular}%
}
\\[6pt] 
  \raggedright
\small{$1$ S:SkyScapes \cite{azimi2019skyscapes}, R:RSVGD \cite{sun2022visual}, DI:DIOR \cite{li2020object}, DO:DOTA \cite{xia2018dota}, Vi:VisDrone \cite{du2019visdrone}, L:LoveDA \cite{wang2loveda}, P:Potsdam \cite{bayanlou2021multi}, H:HRRSD \cite{zhang2019hierarchical}, A:AID \cite{xia2017aid}, FM:fMoW \cite{christie2018functional}, C:Chesapeake \cite{robinson2019large}, DG:DeepGlobe \cite{demir2018deepglobe}, FB:Five-Billion-Pixel \cite{tong2023enabling}, FL:FLAIR \cite{garioud2023flair}, GI:GID-15 \cite{GID2020}, GL:Globe230K \cite{shi2023globe230k}, M:MiniFrance \cite{castillo2022semi}, Va:Vaihingen \cite{bayanlou2021multi}, FA:FAIR1M \cite{sun2022fair1m}.}\\
\small{$2$ Numbers in parentheses denote the average text length for the referring, interactive, and reasoning tasks, respectively.}
\end{table*}

\subsection{Large Multimodal Models for RS}
Large multimodal models have recently emerged to advance geospatial understanding \cite{wanggeollava, spex}. RSGPT \cite{hu2025rsgpt} was the first to generate detailed image descriptions in RS via natural language dialogue. GeoChat \cite{kuckreja2024geochat} extended this by supporting region-specific inputs and responding with bounding boxes for precise visual localization. EarthGPT \cite{zhang2024earthgpt} and EarthDial \cite{soni2025earthdial} integrate various multi-sensor RS interpretation tasks within a large language and multimodal framework. SkySenseGPT \cite{luo2024skysensegpt} contributes image-level scene graph generation and relationship reasoning, while LHRS-Bot \cite{muhtar2024lhrs} enhances multi-level vision-language alignment. On the other hand, Geopix \cite{ou2025geopix} and GeoPixel \cite{shabbir2025geopixel} extended multimodal interaction to pixel-level outputs, producing guided segmentation masks. Despite these efforts, most existing geospatial multimodal models either lack pixel-level segmentation capabilities or are limited to segmentation tasks without contextual reasoning. This results in restricted task coverage and insufficient integration of instruction understanding, visual grounding, and pixel-level output. Therefore, there is a growing need for unified frameworks capable of jointly addressing instruction comprehension, high-level reasoning, and fine-grained segmentation in complex geospatial scenes.

\section{The GeoSeg-1M Dataset and GeoSeg-Bench}
\subsection{Data Curation and Filtering}
We curate a large-scale multimodal segmentation dataset GeoSeg-1M consisting of 590K unique images, 117categories, and 1.1M image-mask-instruction triplets. GeoSeg-1M is built by integrating multiple RS datasets with pixel-level annotations, as reported in Tab.~\ref{tab:dataset_comparison} for a full list. All datasets were standardized into a unified annotation format.

Although many RS datasets provide pixel-level annotations, their masks often contain fragmented regions, imprecise boundaries, and inconsistent class labels. Such noise degrades supervision quality. To address this issue, we developed a systematic mask filtering pipeline that refines raw annotations by decomposing masks into connected regions, removing unreliable areas, and automatically evaluating the remaining ones using InternVL3 \cite{zhu2025internvl3} through a designed prompt (see Supplementary Material for details). Only high-quality masks are preserved for subsequent instruction generation.

\begin{figure*}[htbp]
    \centering
    \includegraphics[width=0.9\linewidth]{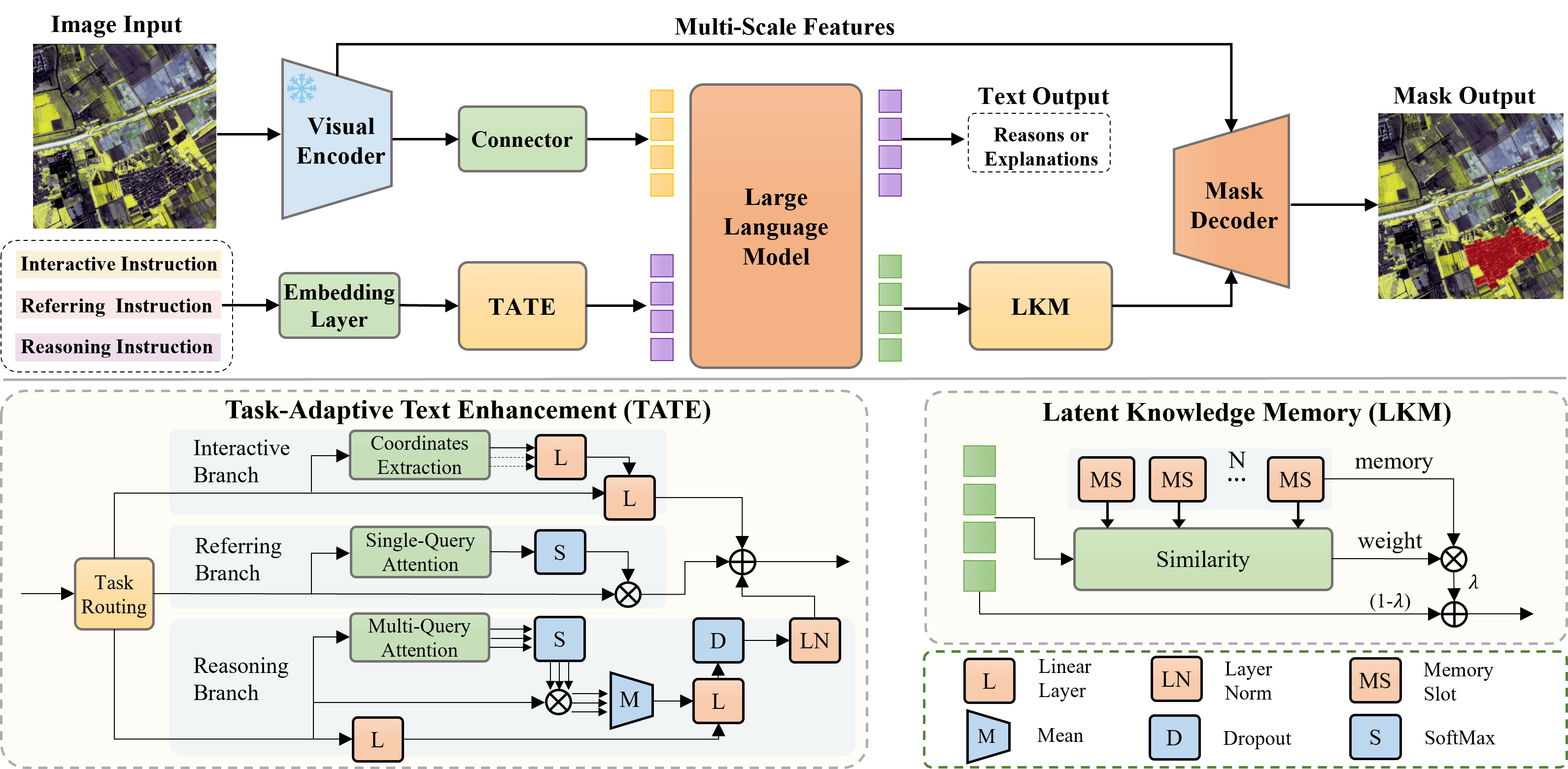}
    \caption{The diagram of UniGeoSeg. The top indicates the whole pipeline, and the bottom describes each module.}
    \label{fig:method}
\end{figure*}

\subsection{Instruction Generation}
To construct multimodal segmentation instructions, we employ a two-stage framework that utilized high-performance proprietary and open-source vision–language models for automated textual instructions generation and quality control. 
Specifically, instructions are generated by GPT-4o \cite{hurst2024gpt} and subsequently cross-evaluated by open-source models (InternVL3-78B \cite{zhu2025internvl3} and QwenVL2-72B \cite{wang2024qwen2}) to ensure clarity, alignment, and reasoning depth. Task-specific generation pipelines for reasoning, referring, and interactive segmentation are designed as follows.

\textbf{Reasoning segmentation data.} We first select semantically diverse images with rich category coverage. For each masked region that no other region of the same category exists, GPT-4o is prompted with category list of the image to generate a reasoning question focusing on its attributes or functional roles. Such cases are defined as attribute reasoning samples. If one or two additional same-category regions exist, the case is treated as a contextual reasoning sample, with regions highlighted in distinct colors. GPT-4o is then instructed to describe their salient spatial relations and generate a question requiring contextual or relational reasoning. All image–mask–question triplets are cross-evaluated by open-source models, and only high-quality samples are retained. Owing to the stricter semantic filtering, the resulting reasoning dataset contains approximately 105k samples. 

\textbf{Referring segmentation data.} We design specialized prompts that guide GPT-4o to generate referring expressions emphasizing spatial and contextual grounding rather than direct visual mention. This design encourages descriptions involving relative positions, neighboring context, and inter-object spatial relations, thereby enriching the spatial semantics expressed in the dataset.
The generated expressions are then cross-scored to ensure linguistic clarity and visual–textual consistency. The final referring segmentation dataset contains approximately 336k samples.

\textbf{Interactive segmentation data.} Fixed-format textual instructions are generated directly from mask geometry. Bounding boxes are defined as tight enclosing rectangles, and one to three random points are sampled within each mask. These textual prompts simulate point- or box-based user interactions, yielding approximately 481k interactive samples. Comprehensive descriptions of the generation and quality-control pipelines, as well as the prompts used for each stage, are included in the Supplementary Material.

In addition, to further enrich the dataset and improve coverage across diverse segmentation tasks, we incorporate the training splits of existing RS referring and reasoning segmentation datasets, including RRSIS-D\cite{liu2024rotated}, RemoteSAM\cite{yao2025remotesam}, and EarthReason\cite{li2025segearthr1}. While some images from these datasets may overlap with previously annotated samples, each is treated as a distinct training instance due to its unique textual instruction.

\subsection{Dataset Statistics and Analysis}

The GeoSeg-1M comprises 590,413 unique RS images and 1,148,504 image–mask–text triplets, establishing the first million-scale multimodal segmentation dataset in this field. It spans 117 object and land-cover categories after semantic merging, with an upper limit imposed per categories to alleviate long-tail imbalance. The textual instructions are linguistically diverse, with varied length distributions. We report the average text length for each task to reflect the complexity of the instructions. Beyond textual diversity, GeoSeg-1M also covers varied spatial resolutions and scene scales, offering the widest spatial coverage among existing multimodal datasets. Tab. \ref{tab:dataset_comparison} compares these characteristics, underscoring the superiority of GeoSeg-1M in scale, language complexity, and spatial diversity for training unified multimodal segmentation models.

\subsection{Benchmark}
We carefully curate GeoSeg-Bench, a benchmark building upon GeoSeg-1M with strict patch-level deduplication, which containes 2,870 interactive, 2,311 referring, and 1,711 reasoning segmentation samples.
Each sample is manually cross-validated by two domain experts, ensuring high-quality image–mask–instruction alignment and faithfully reflecting the multimodal understanding challenges in geospatial imagery.

GeoSeg-Bench is designed to evaluate a model’s capability for fine-grained instruction grounding, spatial understanding, and cross-modal reasoning across three segmentation paradigms. We benchmark a series of multimodal segmentation models, including general-purpose models~\cite{lai2024lisa,ren2024pixellm,zhang2024psalm} and RS-oriented models~\cite{ou2025geopix,shabbir2025geopixel,yao2025remotesam,shu2025earthmind,quenumlisat,li2025segearthr1}. For fairness, several models are further fine-tuned on GeoSeg-1M to assess their adaptability.

Results in Tab.~\ref{tab:metrics_comparison} show that existing models perform poorly without fine-tuning. While fine-tuning leads to substantial improvements, reasoning-task performance remains relatively weak. These findings underscore the need for a unified, reasoning-aware framework for multimodal segmentation in RS. Further analysis can be found in Section~\ref{sec:exp_analysis}.

\section{A Strong Baseline: UniGeoSeg}
\subsection{Overview}
To further advance multi-task instruction-driven multimodal segmentation, we propose UniGeoSeg, a unified vision–language segmentation model that performs reasoning, referring, and interactive segmentation within a single framework. UniGeoSeg comprises three main components: a hierarchical visual encoder that extracts multi-scale spatial features, an LLM that interprets textual instructions, and a pixel-level decoder that generates segmentation masks.

Based upon this core architecture, we introduce three key mechanisms: task-adaptive text enhancement (TATE), latent knowledge memory (LKM), and progressive task scheduling (PTS). These designs jointly enable UniGeoSeg to unify diverse segmentation paradigms while maintaining contextual coherence in RS imagery. The overview of UniGeoSeg is shown as Fig.~\ref{fig:method}.

\subsection{Task-Adaptive Text Enhancement}
To enable a unified text encoder to effectively handle heterogeneous segmentation instructions, we introduce the TATE mechanism. The key insight is that instructions across different segmentation paradigms entail distinct levels of semantic focus and alignment with visual content.

For interactive segmentation, we explicitly encode user-provided spatial cues (\eg, click points or bounding boxes) into the textual embedding space of the LLM. Let $\mathbf{E}_t$ denote the textual embeddings obtained from the embedding layer after tokenization, and $\mathbf{C}_t$ the set of spatial coordinates provided by the user. When multiple coordinates are provided, they are first concatenated and then projected into the same dimensional space as $\mathbf{E}_t$ via a linear layer. A subsequent linear fusion layer adaptively integrates the coordinate-aware features with the text embeddings:
\begin{equation}
    \tilde{\mathbf{E}}_{\text{int}} = \mathrm{Fusion}\left(\mathbf{E}_t, \mathbf{Proj}(\mathbf{C}_t)\right).
\end{equation}
This enhancement explicitly injects and further emphasizes the coordinate-aware spatial cues into the text embedding space, enabling segmentation to be effectively conditioned on user-provided guidance.

For referring segmentation, we enhance token embeddings through an attention mechanism using a single learnable query vector $\mathbf{q}$:
\begin{equation}
\tilde{\mathbf{E}}_{\text{ref}} = \mathrm{softmax}\Big(\frac{\mathbf{q} \cdot \mathbf{E}_t^\top}{\sqrt{d}}\Big) \mathbf{E}_t.
\end{equation}
This query computes similarity with each token embedding in $\mathbf{E}_t$, producing attention weights that selectively emphasize keywords and object-relevant cues.

For reasoning segmentation, the model needs to capture diverse semantic relations and contextual dependencies, such as spatial relations, attribute constraints, and causal reasoning cues \cite{li2025segearthr1}.
To achieve this, we first apply multi-query attention over the token embeddings $\mathbf{E}_t$, where each query learns to attend to a specific reasoning dimension.
In parallel, a linear layer aggregates global information. The outputs from these two streams are fused to obtain the reasoning-enhanced embeddings:
\begin{equation}
\mathbf{E}_{\text{res}} = \frac{1}{h}\sum_{i=1}^{h} \mathrm{softmax}\Big(\frac{\mathbf{q}_i \cdot \mathbf{E}_t^\top}{\sqrt{d}}\Big) \mathbf{E}_t + \mathbf{E}_t \mathbf{W}_G,
\end{equation}
where $h$ is the number of learnable queries and $\mathbf{W}_G$ is a learnable weight matrix for global aggregation.
Finally, dropout and layer normalization are applied to enhance stability when processing long text sequences:
\begin{equation}
\tilde{\mathbf{E}}_{\text{res}} = \mathrm{LayerNorm}(\mathrm{Dropout}(\mathbf{E}_{\text{res}})).
\end{equation}
This design enables the model to synthesize multi-faceted reasoning patterns with holistic instruction semantics, while maintaining robustness and stability for long-range reasoning in segmentation. 

After obtaining the task-specific enhanced embeddings, the final embedding $\tilde{\mathbf{E}}$ sent into the language model is selected based on the segmentation task.
\begin{equation}
    \tilde{\mathbf{E}} =
    \begin{cases}
        \tilde{\mathbf{E}}_{\text{int}}, & \text{if interactive segmentation},\\
        \tilde{\mathbf{E}}_{\text{ref}}, & \text{if referring segmentation},\\
        \tilde{\mathbf{E}}_{\text{res}}, & \text{if reasoning segmentation}.
    \end{cases}
\end{equation}
This ensures that the LLM receives the appropriately enhanced embeddings corresponding to the specific task, enabling effective task-adaptive guidance for segmentation.

By designing separate enhancement pathways for different task paradigms, TATE provides a flexible and modular framework. Each pathway is lightweight yet effective, allowing the model to handle heterogeneous instructions without introducing significant computational overhead.

\subsection{Latent Knowledge Memory}
Although multi-task training allows the model to handle diverse instruction-driven segmentation tasks, each task often learns isolated representations, leading to limited semantic transfer across task boundaries.
To encourage cross-task knowledge sharing, we introduce the LKM module consisting of $N$ learnable memory slots $\{\mathbf{M}_n\}_{n=1}^N$ storing latent, task-agnostic representations distilled from previous instruction–mask pairs.

Let $\mathbf{H} \in \mathbb{R}^{L \times d}$ denote the sequence of $L$ embeddings produced by LLM, where $d$ is the hidden state dimension. We compute attention-based similarity between $\mathbf{H}$ and each memory slot to retrieve latent knowledge as follows:
\begin{equation}
\mathbf{Z} = \sum_{n=1}^N \mathrm{softmax}\big(\mathbf{H}\,\mathbf{M}_n^\top\big) \mathbf{M}_n,
\end{equation}
where $\mathbf{Z} \in \mathbb{R}^{L \times d}$ is the aggregated latent knowledge.

To avoid excessive reliance on latent priors, we fuse the retrieved knowledge with the original embeddings through a weighting scheme:
\begin{equation}
\tilde{\mathbf{H}} = (1 - \lambda)\,\mathbf{H} + \lambda\,\mathbf{Z},
\end{equation}
where $\lambda$ is a hyperparameter that controls the influence of memory. The fused representation $\tilde{\mathbf{H}}$ is then passed to the mask decoder along with multi-scale visual features to generate the output mask.

\subsection{Progressive Task Scheduling}

In multi-task segmentation, the three tasks exhibit different levels of difficulty and data volume. Interactive segmentation is relatively easy and abundant, requiring mainly spatial understanding; referring segmentation is moderately challenging, demanding contextual reasoning and object localization; reasoning segmentation is the most difficult, involving long text, global context integration, and external knowledge, while high-quality samples are scarce.

To balance these differences, we adopt the PTS strategy. 
During training, the proportion of interactive samples is gradually reduced, referring samples are kept stable, and reasoning samples are progressively increased dynamically.
Early emphasis on interactive tasks helps the model acquire basic spatial reasoning, while later reduction prevents overfitting and encourages focus on harder reasoning tasks, improving open-world generalization. Similar to curriculum learning~\cite{bengio2009curriculum}, the PTS effectively balances differences in task difficulty and data volume, enabling a smooth transition from easy to difficult tasks.

\section{Experiments}
\label{sec:exp_analysis}

\subsection{Experimental Setup}
\textbf{Datasets and tasks.} 
We train our baseline model on the GeoSeg-1M dataset. The model is then evaluated on GeoSeg-Bench as well as the validation and test sets of RRSIS-D and EarthReason, demonstrating its ability to handle diverse tasks in a single framework. To evaluate zero-shot generalization, we further train our model from scratch on a DIOR-excluded subset of GeoSeg-1M. The resulting model is tested on 4,000 samples from SIOR \cite{wang2023samrs} for interactive segmentation, with half of the samples using point prompts and half using box prompts. Additionally, we assess zero-shot visual localization on the DIOR-RSVG \cite{zhan2023rsvg} test set, where the target regions are represented by the tightest enclosing rectangles of the masks.

\begin{table}
  \caption{Results on the GeoSeg-Bench. The best results are shown in bold, second-best results are underlined.}
  \label{tab:metrics_comparison}
  \centering
  \resizebox{\columnwidth}{!}{
  \begin{tabular}{@{}lcccccc}

    \toprule
    \multirow{2}{*}{Method} & \multicolumn{2}{c}{Interactive} & \multicolumn{2}{c}{Referring} & \multicolumn{2}{c}{Reasoning} \\
    \cmidrule(lr){2-3} \cmidrule(lr){4-5} \cmidrule(lr){6-7}
    & cIoU & gIoU & cIoU & gIoU & cIoU & gIoU \\
    \midrule
    LISA \cite{lai2024lisa}  & 2.52 & 3.12 & 3.53 & 4.56 & 7.09 & 5.77 \\
    PixelLM \cite{ren2024pixellm} & 0.08 & 0.11 & 6.04 & 5.80 & 6.36 & 6.30 \\
    PSALM \cite{zhang2024psalm} & 6.35 & 10.83  & 31.77 & 18.91 & 11.88 & 9.27 \\
    HIPIE \cite{wang2023hierarchical} & 6.01 & 12.06 & 29.25 & 39.14 & 7.90 & 11.76 \\
    SegLLM \cite{wangsegllm} & 8.97 & 17.72 & 15.90 & 35.27 & 11.06 & 16.41 \\
    Geopixel \cite{shabbir2025geopixel} & 17.21 & 18.71  & 37.34 & 40.14 & 27.36 & 26.71 \\
    Geopix \cite{ou2025geopix}  & 15.28 & 17.63 & 28.81 & 28.52 & 20.80 & 18.25 \\
    RemoteSAM \cite{yao2025remotesam}  & 4.85 & 6.49 & 12.06 & 29.31 & 8.09 & 9.01 \\
    Earthmind \cite{shu2025earthmind}  & 16.38 & 16.57 & 44.53 & 46.48 & 31.01 & 28.80 \\
    LISAT \cite{quenumlisat}  & 4.27 & 5.52 & 37.87 & 40.14 & 22.06 & 20.09 \\
    Segearth-R1 \cite{li2025segearthr1}  & 4.88 & 5.28 & 14.65 & 9.39 & 12.72 & 11.97 \\
    \cmidrule(lr){1-7}
    \multicolumn{7}{l}{\textit{Finetuned on GeoSeg-1M}} \\
    \cmidrule(lr){1-7}
    PSALM  & 70.78 & 74.10 & 68.70 & 71.15 & 47.53 & 49.59 \\
    Geopixel  & 42.62 & 46.48 & 42.70 & 45.50 & 30.25 & 28.51 \\
    Earthmnind  & 67.74 & 70.89 & 48.09 & 49.24 & 36.84 & 25.71 \\
    LISAT  & 68.43 & 73.00 & 59.82 & 62.46 & 41.53 & 31.25 \\
    Segearth-R1  & \underline{72.09} & \underline{75.00} & \underline{70.76} & \underline{72.98} & \underline{53.31} & \underline{51.56} \\
    UniGeoSeg (ours)  & \textbf{74.44} & \textbf{75.56} & \textbf{72.93} & \textbf{74.58} & \textbf{58.35} & \textbf{53.12} \\
    \bottomrule
  \end{tabular}
  }
\end{table}

\begin{figure*}[t]
    \centering
    \includegraphics[width=0.75\linewidth]{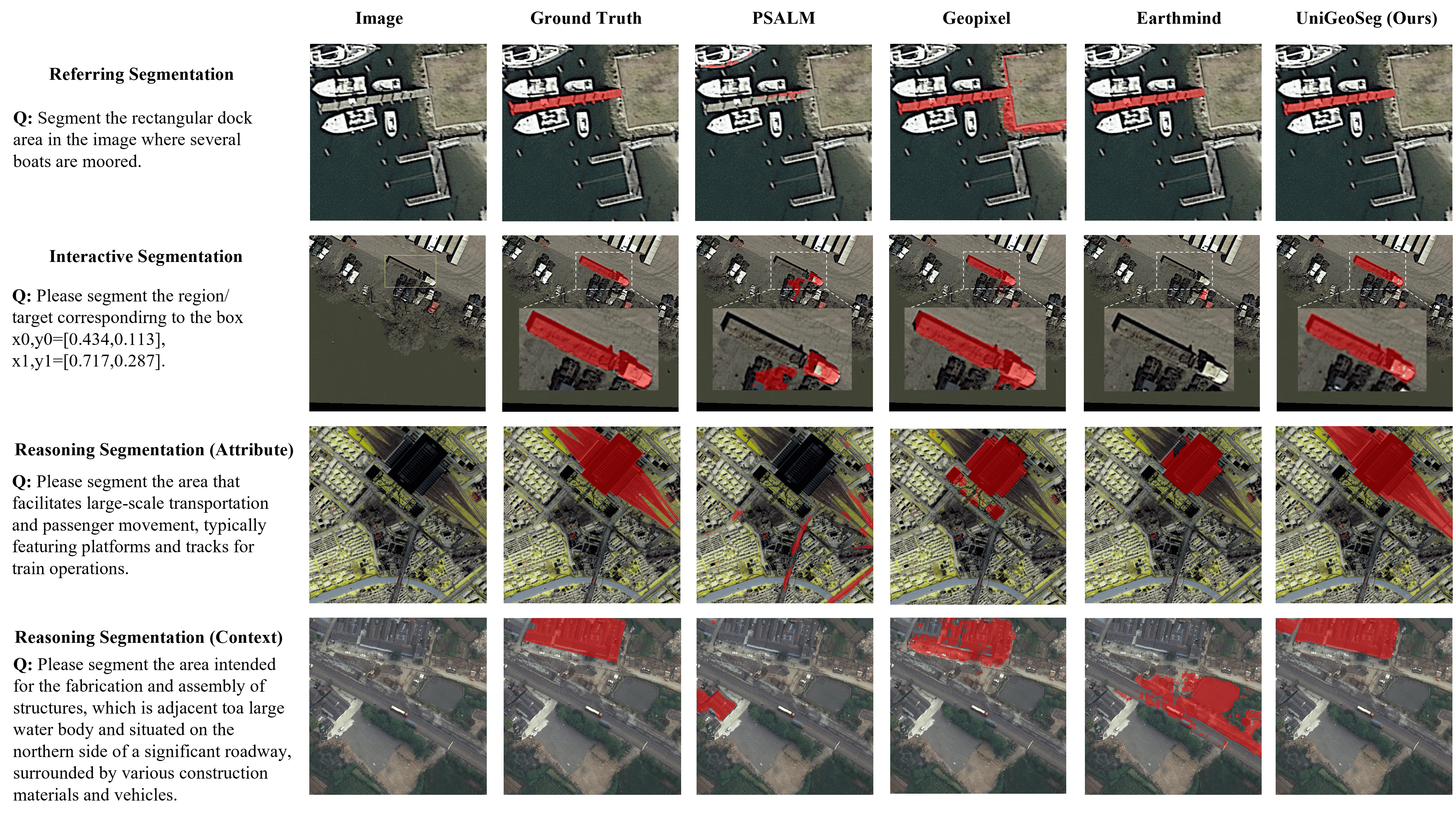}
    \caption{Qualitative examples of the segmentations generated by UniGeoSeg and comparative methods on GeoSeg-Bench.}
    \label{fig:seg}
\end{figure*}

\noindent \textbf{Evaluation metrics.}
Following standard referring segmentation benchmarks (\eg, RefCOCO \cite{yu2016modeling}), we report both global IoU (gIoU) and cumulative IoU (cIoU) for segmentation evaluation, and adopt gIoU as the primary metric, as it equally weights each sample and better captures per-instance performance on multi-scale and irregular targets typical in RS. The cIoU, computed over all pixels, is reported as a complementary area-weighted metric to reflect overall segmentation quality.

The gIoU measures per-sample mask accuracy and is formulated as:
\begin{equation}
\text{gIoU} = \frac{1}{N} \sum_{i=1}^{N} 
\frac{|M_i^{\text{pred}} \cap M_i^{\text{gt}}|}{|M_i^{\text{pred}} \cup M_i^{\text{gt}}|},
\end{equation}
where $M_i^{\text{pred}}$ and $M_i^{\text{gt}}$ denote the predicted and ground-truth masks for sample $i$. 
The cIoU is computed over all pixels across the samples:
\begin{equation}
\text{cIoU} = 
\frac{\sum_i |M_i^{\text{pred}} \cap M_i^{\text{gt}}|}
{\sum_i |M_i^{\text{pred}} \cup M_i^{\text{gt}}|}.
\end{equation}
For visual grounding experiments, we compute the gIoU and cIoU of the predicted bounding boxes with respect to the ground-truth boxes.

\noindent\textbf{Implementation details.} 
Our model adopts Phi-1.5 \cite{li2023phi15} as the language backbone, Swin-B \cite{liu2021swin} as the visual encoder, and Mask2Former \cite{cheng2022masked} as the segmentation decoder. The model is trained with bfloat16 precision. The visual encoder and decoder are initialized from pretrained Mask2Former weights. All input images are resized to $512 \times 512$ for training and evaluation. We employ the AdamW \cite{loshchilov2017decoupled} optimizer with an initial learning rate of $1\times10^{-4}$ and a cosine decay schedule. The visual encoder is frozen during training. The batch size is set to 16, and each experiment is trained for 3 epochs. We apply the proposed PTS strategy where the sampling weight of interactive segmentation gradually decreases to $0.7$, with the remaining weight allocated to reasoning samples. All experiments are conducted on eight NVIDIA A800 GPUs.

\subsection{Main Results}

\textbf{Results on GeoSeg-Bench.} We first evaluate a wide range of models, including general-purpose LMMs and RS-specific methods, on the interactive, referring, and reasoning tasks of GeoSeg-Bench. As shown in Tab.~\ref{tab:metrics_comparison}, our model achieves state-of-the-art performance on all three tasks, outperforming existing general and RS large multimodal models. Notably, other models perform poorly when directly tested on  GeoSeg-Bench without training, indicating their limited capability in understanding context and performing complex reasoning in geospatial scenes. We further fine-tune some well-performing models on our dataset. While fine-tuning significantly improves their results, they still lag behind our model, demonstrating the superiority of our model in capturing spatial grounding, multimodal context, and reasoning. Some examples of the segmentations generated by UniGeoSeg and comparative methods on GeoSeg-Bench are shown in Fig.~\ref{fig:seg}.

\noindent\textbf{Results on EarthReason and RRSIS-D.} We further evaluate a variety of methods, including our model, on EarthReason and RRSIS-D, corresponding to reasoning and referring segmentation, respectively. As shown in Tab.~\ref{tab:combined_results}, our model maintains superior performance across both tasks. It achieves absolute gains of 6.65 and 2.16 percentage points in cIoU and gIoU on EarthReason test set over the previous best. The consistent performance across EarthReason and RRSIS-D highlights the general applicability of UniGeoSeg to RS referring and reasoning segmentation.

\begin{table}
  \caption{Results on EarthReason and RRSIS-D. The best results are shown in bold, second-best results are underlined.}
  \label{tab:combined_results}
  \centering
  \resizebox{\columnwidth}{!}{
  \begin{tabular}{@{}lcccc|cc@{}}
    \toprule
    \multirow{2}{*}{Method} 
      & \multicolumn{2}{c}{EarthReason (Val)} 
      & \multicolumn{2}{c|}{EarthReason (Test)} 
      & \multicolumn{2}{c}{RRSIS-D} \\
    \cmidrule(lr){2-3} \cmidrule(lr){4-5} \cmidrule(lr){6-7}
      & cIoU & gIoU & cIoU & gIoU & cIoU & gIoU \\
    \midrule
    LISA \cite{lai2024lisa} & 57.39 & 59.10 & 61.04 & 60.88 & 27.84 & 26.78 \\
    PixelLM \cite{ren2024pixellm} & 57.79 & 59.22 & 57.94 & 60.01 & 33.89 & 31.65 \\
    PSALM \cite{zhang2024psalm} & 62.03 & 64.61 & 66.61 & 68.30 & - & - \\
    NExT-Chat \cite{zhang2023nextchatlmmchatdetection} & - & - & - & - & 26.98 & 24.98 \\
    Geoground \cite{zhou2025geogroundunifiedlargevisionlanguage} & - & - & - & - & 61.10 & 60.50 \\
    Geopixel \cite{shabbir2025geopixel} & 54.23 & 52.13 & 53.90 & 52.53 & \textbf{81.77} & 67.99 \\
    Segearth-R1 \cite{li2025segearthr1} & 64.13 & 68.25 & 68.60 & 70.75 & 67.56 & 66.40 \\
    RemoteReasoner \cite{yao2025remotereasoner} & \underline{67.80} & \underline{69.13} & \underline{69.02} & \underline{70.96} & 54.29 & 50.97 \\
    UniGeoSeg (ours) & \textbf{73.30} & \textbf{72.54} & \textbf{74.61} & \textbf{73.08} & \underline{79.78} & \textbf{69.25} \\
    \bottomrule
  \end{tabular}
  }
\end{table}

\noindent\textbf{Zero-shot results.} We subsequently report the zero-shot results on interactive segmentation and visual grounding tasks, as shown in Tab.~\ref{tab:interactive_segmentation_results} and Tab.~\ref{tab:zero_shot_rsvg_dior}, respectively. The model achieves competitive performance in both settings, demonstrating strong spatial modeling and consistently robust generalization across RS imagery.

\begin{table}
  \caption{Results of zero-shot interactive segmentation on SIOR. The best results are bold, second-best results are underlined.}
  \label{tab:interactive_segmentation_results}
  \centering
  \resizebox{\columnwidth}{!}{
  \begin{tabular}{@{}lcccc|cc@{}}
    \toprule
     \multirow{2}{*}{Method} & \multicolumn{2}{c}{Box} & \multicolumn{2}{c|}{Point} & \multicolumn{2}{c}{Average}\\
     \cmidrule(lr){2-3} \cmidrule(lr){4-5} \cmidrule(lr){6-7}
    & cIoU & gIoU & cIoU & gIoU & cIoU & gIoU\\
    \midrule
    SAM2 \cite{ravi2024sam2} & \textbf{91.23} & \textbf{92.27} & \underline{71.47} & \underline{76.45} & \underline{80.34} & \underline{84.36} \\
    HQ-SAM \cite{sam_hq} & \underline{91.91} & 90.81 & 69.25 & 74.63 & 79.12 & 83.27 \\
    RemoteSAM \cite{yao2025remotesam} & 12.02 & 21.89 & 12.32 & 20.45 & 12.18 & 21.17 \\
    PSALM \cite{zhang2024psalm} & 48.09 & 57.36 & 40.60 & 51.08 & 44.01 & 51.95 \\
    UniGeoSeg (ours) & 90.60 & \underline{90.94} & \textbf{84.61} & \textbf{86.60} & \textbf{87.42} & \textbf{88.77} \\
    \bottomrule
  \end{tabular}
  }
\end{table}

\begin{table}
  \caption{Results of zero-shot vision grounding on RSVG-DIOR test set, best results are in bold, second-best results are underlined.}
  \label{tab:zero_shot_rsvg_dior}
  \centering
    \resizebox{\columnwidth}{!}{
  \begin{tabular}{@{}lccccc@{}}
    \toprule
    Method & cIoU & gIoU & PR@0.3 & PR@0.5 & PR@0.7 \\
    \midrule
    OV-VG \cite{wang2024OV-VG} & 18.34 & 22.07 & 27.78 & 19.16 & 12.93 \\
    GroundVLP \cite{shen2024groundvlp} & 17.51 & 17.14 & 21.66 & 16.83 & 12.27 \\
    DiffSegmenter \cite{DiffSegmenter} & 26.92 & 28.47 & 39.80 & 24.29 & 13.01\\
    DiffPNG \cite{DiffPNG} & 25.65 & 26.93 & 37.92 & 23.18 & 11.35 \\
    LISA \cite{lai2024lisa} & - & 27.78 & - & 25.80 & -\\
    GeoChat \cite{kuckreja2024geochat} & - & 32.53 & - & 27.61 & -\\
    Qwen2.5-VL \cite{wang2024qwen2} & - & 31.93 & - & 30.35 & -\\
    RSVG-ZeroOV \cite{RSVG-ZeroOV} & \underline{31.28} & \underline{34.49} & \underline{45.71} & \underline{31.39} & \underline{16.74}\\
    UniGeoSeg (ours) & \textbf{70.82} & \textbf{59.67} & \textbf{74.94} & \textbf{67.84} & \textbf{55.11} \\
    \bottomrule
  \end{tabular}
  }
\end{table}

\subsection{Ablations}
All ablation experiments are conducted on a representative subset of GeoSeg-1M with DIOR-derived data excluded.
We report the performance in terms of gIoU on the referring and reasoning tasks of GepSeg-Bench, as well as on interactive segmentation task of SIOR, to systematically quantify the effect of each component.

\begin{table}[t]
  \caption{Ablation study of TATE and LKM modules. \checkmark indicates the module is enabled.}
  \label{tab:ablation_modules}
  \centering
  \resizebox{\columnwidth}{!}{
  \begin{tabular}{@{}cc|ccc@{}}
    \toprule
     TATE & LKM & Interactive & Referring & Reasoning \\
    \midrule
     &   & 82.51 & 64.62 & 32.88 \\
    \checkmark &   & 84.61 (+2.10) & 64.70 (+0.08) & 35.64 (+2.76) \\
     & \checkmark & 81.97 (-0.54) & 65.21 (+0.59) & 33.85 (+0.97) \\
    \checkmark & \checkmark & \textbf{84.84} \textbf{(+2.33)} & \textbf{66.37} \textbf{(+1.75)} & \textbf{37.06} \textbf{(+4.18)} \\
    \bottomrule
  \end{tabular}
  }
\end{table}

\noindent\textbf{TATE and LKM.} We first perform ablation studies on the TATE and LKM modules. By removing these modules individually, we examine their contributions to the overall model performance. As shown in Tab.~\ref{tab:ablation_modules}, the removal of either module leads to a noticeable drop in performance across tasks, confirming the effectiveness of TATE for multi-task text disentanglement and spatial reasoning, and of LKM for cross-task knowledge sharing.

\begin{table}
  \small
  \caption{Ablation study of the branches in TATE.}
  \label{tab:tate_ablation}
  \centering
  \resizebox{\columnwidth}{!}{
  \begin{tabular}{@{}lccc@{}}
    \toprule
     & Interactive & Referring & Reasoning \\
    \midrule
    w/o TATE & 81.97 & 65.21 & 33.85 \\
    All referring branch & 81.29 (-0.68) & 65.44 (+0.23) & 33.88 (+0.03) \\
    All reasoning branch & 83.16 (+1.19) & 65.73 (+0.52) & 36.01 (+2.16) \\
    TATE & \textbf{84.84} (\textbf{+2.87}) & \textbf{66.37} (\textbf{+1.16}) & \textbf{37.06} (\textbf{+3.21}) \\
    \bottomrule
  \end{tabular}
  }
\end{table}

\noindent\textbf{Branches in TATE.} We further conduct an internal ablation study of the TATE module. Specifically, we report the performance when three tasks use the same referring branch or reasoning branch without task-specific customization. As reported in Tab.~\ref{tab:tate_ablation}, applying the same mechanism across all branches results in lower gIoU on the three tasks, highlighting the effectiveness of assigning lightweight, task-specific enhancements within TATE to better capture the distinct textual reasoning requirements of each task.

\begin{table}[t]
  \small
  \caption{Ablation on the memory slot size ($N$) and fusion weight ($\lambda$) of the LKM module. Performance is reported as gIoU on Referring / Reasoning tasks. The best results are shown in bold.}
  \label{tab:ablation_hyperparams}
  \centering
  \begin{tabular}{@{}lcc@{}}
    \toprule
     & $N=4$ & $N=8$ \\
    \midrule
    $\lambda=0.0$ & 64.70 / 35.64 & 64.70 / 35.64 \\
    $\lambda=0.1$ & 65.79 / 36.23 & 65.14 / 35.81 \\
    $\lambda=0.2$ & \textbf{66.37 / 37.06} & 66.17 / 36.95 \\
    $\lambda=0.3$ & 65.73 / 35.47 & 66.06 / 36.72 \\
    \bottomrule
  \end{tabular}
\end{table}

\begin{table}[t]
  \small
  \caption{Ablation of different sampling strategies.}
  \label{tab:sampling_strategies}
  \centering
    \resizebox{\columnwidth}{!}{
  \begin{tabular}{@{}lccc@{}}
    \toprule
    Strategy & Interactive & Referring & Reasoning \\
    \midrule
    Average sampling & 88.76 & 73.14 & 51.11 \\
    PTS & \textbf{88.77} (\textbf{+0.01}) & \textbf{73.21} (\textbf{+0.07}) & \textbf{51.42} (\textbf{+0.31}) \\
    \bottomrule
  \end{tabular}
  }
\end{table}

\noindent\textbf{Hyperparameters in LKM.} We also perform an ablation study on the LKM module, which involves two key hyperparameters: the number of memory slots ($N$) and the fusion weight factor ($\lambda$). By systematically varying these parameters and evaluating the model on our benchmark and the DIOR-based interactive segmentation task, we find that $N=4$ and $\lambda=0.2$ yield the best performance, as shown in Tab.~\ref{tab:ablation_hyperparams}, providing a practical setting for balancing memory capacity and integration strength.

\noindent\textbf{PTS strategy.} Finally, we conduct an ablation study on the PTS training strategy using the subset of the GeoSeg-1M with all DIOR-derived data excluded. Compared to uniform sampling, PTS strategy improves performance on both the referring expression and reasoning segmentation tasks without compromising interactive segmentation, as shown in Tab.~\ref{tab:sampling_strategies}. These results validate the effectiveness of the PTS strategy in balancing multiple tasks during training.

\section{Conclusion}
We present GeoSeg-1M, the first million-scale multimodal segmentation dataset for RS, together with GeoSeg-Bench, a comprehensive benchmark designed to evaluate spatial reasoning, contextual understanding, and multi-task performance in complex geospatial scenes.
Building upon these resources, we introduce UniGeoSeg, a unified baseline for instruction-driven multimodal segmentation that integrates multi-task alignment and long-text reasoning within a single framework.
Extensive experiments demonstrate that UniGeoSeg achieves state-of-the-art performance on both GeoSeg-Bench and multiple public datasets, while exhibiting strong zero-shot generalization on interactive segmentation and visual grounding tasks.
Overall, our unified approach advances reasoning-driven multimodal understanding in RS and establishes a scalable foundation for future research in open-world geospatial intelligence.

\section*{Acknowledgments}
This work was supported in part by the New Generation Artificial Intelligence-National Science and Technology Major Project (No. 2025ZD0123602), the Fundamental and Interdisciplinary Disciplines Breakthrough Plan of the Ministry of Education of China (No. JYB2025XDXM101), the National Natural Science Foundation of China (No. 62225113, No. 62501050, No. 624B2109, No. 623B2079), the Zhongguancun Academy Project (No. 20240308), and the Key Technology Research Project of China National Petroleum Corporation (No. 2025ZG82).
{
    \small
    \bibliographystyle{ieeenat_fullname}
    \bibliography{main}

@String(CVPR= {IEEE Conf. Comput. Vis. Pattern Recog.})

@String(ICCV= {Int. Conf. Comput. Vis.})

@String(ECCV= {Eur. Conf. Comput. Vis.})

@String(ICLR = {Int. Conf. Learn. Represent.})

@String(AAAI = {AAAI})

@String(CVPR  = {CVPR})

@String(ICCV  = {ICCV})

@String(ECCV  = {ECCV})

@String(ICLR  = {ICLR})

@inproceedings{gao2024enrich,
title={Enrich Distill and Fuse: Generalized Few-Shot Semantic Segmentation in Remote Sensing Leveraging Foundation Model's Assistance},
author={Gao, Tianyi and Ao, Wei and Wang, Xing-ao and Zhao, Yuanhao and Ma, Ping and Xie, Mengjie and Fu, Hang and Ren, Jinchang and Gao, Zhi},
booktitle={CVPR},
pages={2771--2780},
year={2024}
}

@article{savva2025advances,
  title={Advances in Remote Sensing and AI for Vegetation Monitoring in Power Line Corridors: A Review and Future Directions: A review and future directions},
  author={Savva, Antonis and Kyrkou, Christos and Kolios, Panayiotis and Theocharides, Theocharis},
  journal={IEEE Geoscience and Remote Sensing Magazine},
  year={2025},
  publisher={IEEE}
}

@inproceedings{wang2025xlrs,
  title={Xlrs-bench: Could your multimodal llms understand extremely large ultra-high-resolution remote sensing imagery?},
  author={Wang, Fengxiang and Wang, Hongzhen and Guo, Zonghao and Wang, Di and Wang, Yulin and Chen, Mingshuo and Ma, Qiang and Lan, Long and Yang, Wenjing and Zhang, Jing and others},
  booktitle={CVPR},
  pages={14325--14336},
  year={2025}
}

@inproceedings{ahn2025generalizable,
  title={Generalizable disaster damage assessment via change detection with vision foundation model},
  author={Ahn, Kyeongjin and Han, Sungwon and Park, Sungwon and Kim, Jihee and Park, Sangyoon and Cha, Meeyoung},
  booktitle={AAAI},
  volume={39},
  number={27},
  pages={27784--27792},
  year={2025}
}

@article{TiMo,
      title={TiMo: Spatiotemporal Foundation Model for Satellite Image Time Series}, 
      author={Xiaolei Qin and Di Wang and Jing Zhang and Fengxiang Wang and Xin Su and Bo Du and Liangpei Zhang},
      journal={arXiv preprint arXiv:2505.08723},
      year={2025}
}

@inproceedings{li2025segearth,
  title={Segearth-ov: Towards training-free open-vocabulary segmentation for remote sensing images},
  author={Li, Kaiyu and Liu, Ruixun and Cao, Xiangyong and Bai, Xueru and Zhou, Feng and Meng, Deyu and Wang, Zhi},
  booktitle={CVPR},
  pages={10545--10556},
  year={2025}
}

@inproceedings{liu2024rotated,
  title={Rotated multi-scale interaction network for referring remote sensing image segmentation},
  author={Liu, Sihan and Ma, Yiwei and Zhang, Xiaoqing and Wang, Haowei and Ji, Jiayi and Sun, Xiaoshuai and Ji, Rongrong},
  booktitle={CVPR},
  pages={26658--26668},
  year={2024}
}

@article{li2025segearthr1,
  title={Segearth-r1: Geospatial pixel reasoning via large language model},
  author={Li, Kaiyu and Xin, Zepeng and Pang, Li and Pang, Chao and Deng, Yupeng and Yao, Jing and Xia, Guisong and Meng, Deyu and Wang, Zhi and Cao, Xiangyong},
  journal={arXiv preprint arXiv:2504.09644},
  year={2025}
}

@inproceedings{kirillov2023segment,
  title={Segment anything},
  author={Kirillov, Alexander and Mintun, Eric and Ravi, Nikhila and Mao, Hanzi and Rolland, Chloe and Gustafson, Laura and Xiao, Tete and Whitehead, Spencer and Berg, Alexander C and Lo, Wan-Yen and others},
  booktitle={ICCV},
  pages={4015--4026},
  year={2023}
}

@ARTICLE{mtp,
  author={Wang, Di and Zhang, Jing and Xu, Minqiang and Liu, Lin and Wang, Dongsheng and Gao, Erzhong and Han, Chengxi and Guo, Haonan and Du, Bo and Tao, Dacheng and Zhang, Liangpei},
  journal={IEEE Journal of Selected Topics in Applied Earth Observations and Remote Sensing}, 
  title={MTP: Advancing Remote Sensing Foundation Model via Multitask Pretraining}, 
  year={2024},
  volume={17},
  number={},
  pages={11632-11654}
  }

@article{zhang2024earthmarker,
  title={Earthmarker: A visual prompting multi-modal large language model for remote sensing},
  author={Zhang, Wei and Cai, Miaoxin and Zhang, Tong and Zhuang, Yin and Li, Jun and Mao, Xuerui},
  journal={IEEE Transactions on Geoscience and Remote Sensing},
  year={2024},
  publisher={IEEE}
}

@article{wang2023samrs,
  title={Samrs: Scaling-up remote sensing segmentation dataset with segment anything model},
  author={Wang, Di and Zhang, Jing and Du, Bo and Xu, Minqiang and Liu, Lin and Tao, Dacheng and Zhang, Liangpei},
  journal={NeurIPS},
  volume={36},
  pages={8815--8827},
  year={2023}
}

@inproceedings{chen2025rsrefseg,
  title={RSRefSeg: Referring remote sensing image segmentation with foundation models},
  author={Chen, Keyan and Zhang, Jiafan and Liu, Chenyang and Zou, Zhengxia and Shi, Zhenwei},
  booktitle={IGARSS 2025-2025 IEEE International Geoscience and Remote Sensing Symposium},
  pages={1070--1074},
  year={2025},
  organization={IEEE}
}

@article{lu2025rrsecs,
  title={RRSECS: Referring remote sensing expression comprehension and segmentation},
  author={Lu, Xiaoqiang and Sun, Long and Li, Lingling and Jiao, Licheng and Yang, Yuting and Huang, Zhongjian and Chai, Jinming and Liu, Xu and Liu, Fang and Ma, Wenping and others},
  journal={IEEE Geoscience and Remote Sensing Magazine},
  year={2025},
  publisher={IEEE}
}

@article{ou2025geopix,
  title={GeoPix: A multimodal large language model for pixel-level image understanding in remote sensing},
  author={Ou, Ruizhe and Hu, Yuan and Zhang, Fan and Chen, Jiaxin and Liu, Yu},
  journal={IEEE Geoscience and Remote Sensing Magazine},
  year={2025},
  publisher={IEEE}
}

@inproceedings{pan2025locate,
  title={Locate anything on earth: Advancing open-vocabulary object detection for remote sensing community},
  author={Pan, Jiancheng and Liu, Yanxing and Fu, Yuqian and Ma, Muyuan and Li, Jiahao and Paudel, Danda Pani and Van Gool, Luc and Huang, Xiaomeng},
  booktitle={AAAI},
  volume={39},
  number={6},
  pages={6281--6289},
  year={2025}
}

@inproceedings{shabbir2025geopixel,
  title={GeoPixel: Pixel Grounding Large Multimodal Model in Remote Sensing},
  author={Shabbir, Akashah and Zumri, Mohammed and Bennamoun, Mohammed and Khan, Fahad Shahbaz and Khan, Salman},
  booktitle={ICML},
  year={2025}
}

@article{yao2025remotereasoner,
  title={RemoteReasoner: Towards Unifying Geospatial Reasoning Workflow},
  author={Yao, Liang and Liu, Fan and Lu, Hongbo and Zhang, Chuanyi and Min, Rui and Xu, Shengxiang and Di, Shimin and Peng, Pai},
  journal={arXiv preprint arXiv:2507.19280},
  year={2025}
}

@inproceedings{hu2016segmentation,
  title={Segmentation from natural language expressions},
  author={Hu, Ronghang and Rohrbach, Marcus and Darrell, Trevor},
  booktitle={ECCV},
  pages={108--124},
  year={2016},
  organization={Springer}
}

@inproceedings{li2018referring,
  title={Referring image segmentation via recurrent refinement networks},
  author={Li, Ruiyu and Li, Kaican and Kuo, Yi-Chun and Shu, Michelle and Qi, Xiaojuan and Shen, Xiaoyong and Jia, Jiaya},
  booktitle={CVPR},
  pages={5745--5753},
  year={2018}
}

@inproceedings{shi2018key,
  title={Key-word-aware network for referring expression image segmentation},
  author={Shi, Hengcan and Li, Hongliang and Meng, Fanman and Wu, Qingbo},
  booktitle={ECCV},
  pages={38--54},
  year={2018}
}

@inproceedings{yang2022lavt,
  title={Lavt: Language-aware vision transformer for referring image segmentation},
  author={Yang, Zhao and Wang, Jiaqi and Tang, Yansong and Chen, Kai and Zhao, Hengshuang and Torr, Philip HS},
  booktitle={CVPR},
  pages={18155--18165},
  year={2022}
}

@article{wu2024toward,
  title={Toward robust referring image segmentation},
  author={Wu, Jianzong and Li, Xiangtai and Li, Xia and Ding, Henghui and Tong, Yunhai and Tao, Dacheng},
  journal={IEEE Transactions on Image Processing},
  volume={33},
  pages={1782--1794},
  year={2024},
  publisher={IEEE}
}

@inproceedings{nag2024safari,
  title={SafaRi: Adaptive S equence Tr a ns f ormer for We a kly Supervised R eferring Expression Segmentat i on},
  author={Nag, Sayan and Goswami, Koustava and Karanam, Srikrishna},
  booktitle={ECCV},
  pages={485--503},
  year={2024},
  organization={Springer}
}

@inproceedings{lai2024lisa,
  title={Lisa: Reasoning segmentation via large language model},
  author={Lai, Xin and Tian, Zhuotao and Chen, Yukang and Li, Yanwei and Yuan, Yuhui and Liu, Shu and Jia, Jiaya},
  booktitle={CVPR},
  pages={9579--9589},
  year={2024}
}

@inproceedings{zhang2024psalm,
  title={Psalm: Pixelwise segmentation with large multi-modal model},
  author={Zhang, Zheng and Ma, Yeyao and Zhang, Enming and Bai, Xiang},
  booktitle={ECCV},
  pages={74--91},
  year={2024},
  organization={Springer}
}

@article{yuan2024rrsis,
  title={Rrsis: Referring remote sensing image segmentation},
  author={Yuan, Zhenghang and Mou, Lichao and Hua, Yuansheng and Zhu, Xiao Xiang},
  journal={IEEE Transactions on Geoscience and Remote Sensing},
  volume={62},
  pages={1--12},
  year={2024},
  publisher={IEEE}
}

@article{zhan2023rsvg,
  title={Rsvg: Exploring data and models for visual grounding on remote sensing data},
  author={Zhan, Yang and Xiong, Zhitong and Yuan, Yuan},
  journal={IEEE Transactions on Geoscience and Remote Sensing},
  volume={61},
  pages={1--13},
  year={2023},
  publisher={IEEE}
}

@inproceedings{yao2025remotesam,
  title={Remotesam: Towards segment anything for earth observation},
  author={Yao, Liang and Liu, Fan and Chen, Delong and Zhang, Chuanyi and Wang, Yijun and Chen, Ziyun and Xu, Wei and Di, Shimin and Zheng, Yuhui},
  booktitle={ACM MM},
  pages={3027--3036},
  year={2025}
}

@article{hu2025rsgpt,
  title={Rsgpt: A remote sensing vision language model and benchmark},
  author={Hu, Yuan and Yuan, Jianlong and Wen, Congcong and Lu, Xiaonan and Liu, Yu and Li, Xiang},
  journal={ISPRS Journal of Photogrammetry and Remote Sensing},
  volume={224},
  pages={272--286},
  year={2025},
  publisher={Elsevier}
}

@inproceedings{kuckreja2024geochat,
  title={Geochat: Grounded large vision-language model for remote sensing},
  author={Kuckreja, Kartik and Danish, Muhammad Sohail and Naseer, Muzammal and Das, Abhijit and Khan, Salman and Khan, Fahad Shahbaz},
  booktitle={CVPR},
  pages={27831--27840},
  year={2024}
}

@article{zhang2024earthgpt,
  title={EarthGPT: A universal multimodal large language model for multisensor image comprehension in remote sensing domain},
  author={Zhang, Wei and Cai, Miaoxin and Zhang, Tong and Zhuang, Yin and Mao, Xuerui},
  journal={IEEE Transactions on Geoscience and Remote Sensing},
  volume={62},
  pages={1--20},
  year={2024},
  publisher={IEEE}
}

@inproceedings{soni2025earthdial,
  title={Earthdial: Turning multi-sensory earth observations to interactive dialogues},
  author={Soni, Sagar and Dudhane, Akshay and Debary, Hiyam and Fiaz, Mustansar and Munir, Muhammad Akhtar and Danish, Muhammad Sohail and Fraccaro, Paolo and Watson, Campbell D and Klein, Levente J and Khan, Fahad Shahbaz and others},
  booktitle={CVPR},
  pages={14303--14313},
  year={2025}
}

@article{luo2024skysensegpt,
  title={Skysensegpt: A fine-grained instruction tuning dataset and model for remote sensing vision-language understanding},
  author={Luo, Junwei and Pang, Zhen and Zhang, Yongjun and Wang, Tingzhu and Wang, Linlin and Dang, Bo and Lao, Jiangwei and Wang, Jian and Chen, Jingdong and Tan, Yihua and others},
  journal={arXiv preprint arXiv:2406.10100},
  year={2024}
}

@inproceedings{muhtar2024lhrs,
  title={Lhrs-bot: Empowering remote sensing with vgi-enhanced large multimodal language model},
  author={Muhtar, Dilxat and Li, Zhenshi and Gu, Feng and Zhang, Xueliang and Xiao, Pengfeng},
  booktitle={ECCV},
  pages={440--457},
  year={2024},
  organization={Springer}
}

@article{zhu2025internvl3,
  title={Internvl3: Exploring advanced training and test-time recipes for open-source multimodal models},
  author={Zhu, Jinguo and Wang, Weiyun and Chen, Zhe and Liu, Zhaoyang and Ye, Shenglong and Gu, Lixin and Tian, Hao and Duan, Yuchen and Su, Weijie and Shao, Jie and others},
  journal={arXiv preprint arXiv:2504.10479},
  year={2025}
}

@article{hurst2024gpt,
  title={Gpt-4o system card},
  author={Hurst, Aaron and Lerer, Adam and Goucher, Adam P and Perelman, Adam and Ramesh, Aditya and Clark, Aidan and Ostrow, AJ and Welihinda, Akila and Hayes, Alan and Radford, Alec and others},
  journal={arXiv preprint arXiv:2410.21276},
  year={2024}
}

@article{wang2024qwen2,
  title={Qwen2-vl: Enhancing vision-language model's perception of the world at any resolution},
  author={Wang, Peng and Bai, Shuai and Tan, Sinan and Wang, Shijie and Fan, Zhihao and Bai, Jinze and Chen, Keqin and Liu, Xuejing and Wang, Jialin and Ge, Wenbin and others},
  journal={arXiv preprint arXiv:2409.12191},
  year={2024}
}

@inproceedings{azimi2019skyscapes,
  title={Skyscapes fine-grained semantic understanding of aerial scenes},
  author={Azimi, Seyed Majid and Henry, Corentin and Sommer, Lars and Schumann, Arne and Vig, Eleonora},
  booktitle={ICCV},
  pages={7393--7403},
  year={2019}
}

@inproceedings{sun2022visual,
  title={Visual grounding in remote sensing images},
  author={Sun, Yuxi and Feng, Shanshan and Li, Xutao and Ye, Yunming and Kang, Jian and Huang, Xu},
  booktitle={Proceedings of the 30th ACM International conference on Multimedia},
  pages={404--412},
  year={2022}
}

@article{li2020object,
  title={Object detection in optical remote sensing images: A survey and a new benchmark},
  author={Li, Ke and Wan, Gang and Cheng, Gong and Meng, Liqiu and Han, Junwei},
  journal={ISPRS journal of photogrammetry and remote sensing},
  volume={159},
  pages={296--307},
  year={2020},
  publisher={Elsevier}
}

@inproceedings{xia2018dota,
  title={DOTA: A large-scale dataset for object detection in aerial images},
  author={Xia, Gui-Song and Bai, Xiang and Ding, Jian and Zhu, Zhen and Belongie, Serge and Luo, Jiebo and Datcu, Mihai and Pelillo, Marcello and Zhang, Liangpei},
  booktitle={CVPR},
  pages={3974--3983},
  year={2018}
}

@inproceedings{du2019visdrone,
  title={VisDrone-DET2019: The vision meets drone object detection in image challenge results},
  author={Du, Dawei and Zhu, Pengfei and Wen, Longyin and Bian, Xiao and Lin, Haibin and Hu, Qinghua and Peng, Tao and Zheng, Jiayu and Wang, Xinyao and Zhang, Yue and others},
  booktitle={ICCV workshops},
  pages={0--0},
  year={2019}
}

@inproceedings{wang2loveda,
  title={LoveDA: A Remote Sensing Land-Cover Dataset for Domain Adaptive Semantic Segmentation},
  author={Wang, Junjue and Zheng, Zhuo and Lu, Xiaoyan and Zhong, Yanfei and others},
  booktitle={Thirty-fifth Conference on Neural Information Processing Systems Datasets and Benchmarks Track (Round 2)}
}

@article{bayanlou2021multi,
  title={MULTI-TASK LEARNING FROM FIXED-WING UAV IMAGES FOR 2D/3D CITY MODELLING},
  author={Bayanlou, MR and Khoshboresh-Masouleh, M},
  journal={The International Archives of the Photogrammetry, Remote Sensing and Spatial Information Sciences},
  volume={44},
  pages={1--5},
  year={2021},
  publisher={Copernicus GmbH}
}

@article{zhang2019hierarchical,
  title={Hierarchical and robust convolutional neural network for very high-resolution remote sensing object detection},
  author={Zhang, Yuanlin and Yuan, Yuan and Feng, Yachuang and Lu, Xiaoqiang},
  journal={IEEE Transactions on Geoscience and Remote Sensing},
  volume={57},
  number={8},
  pages={5535--5548},
  year={2019},
  publisher={IEEE}
}

@article{xia2017aid,
  title={AID: A benchmark data set for performance evaluation of aerial scene classification},
  author={Xia, Gui-Song and Hu, Jingwen and Hu, Fan and Shi, Baoguang and Bai, Xiang and Zhong, Yanfei and Zhang, Liangpei and Lu, Xiaoqiang},
  journal={IEEE Transactions on Geoscience and Remote Sensing},
  volume={55},
  number={7},
  pages={3965--3981},
  year={2017},
  publisher={IEEE}
}

@inproceedings{christie2018functional,
  title={Functional map of the world},
  author={Christie, Gordon and Fendley, Neil and Wilson, James and Mukherjee, Ryan},
  booktitle={CVPR},
  pages={6172--6180},
  year={2018}
}

@inproceedings{robinson2019large,
  title={Large scale high-resolution land cover mapping with multi-resolution data},
  author={Robinson, Caleb and Hou, Le and Malkin, Kolya and Soobitsky, Rachel and Czawlytko, Jacob and Dilkina, Bistra and Jojic, Nebojsa},
  booktitle={CVPR},
  pages={12726--12735},
  year={2019}
}

@inproceedings{demir2018deepglobe,
  title={Deepglobe 2018: A challenge to parse the earth through satellite images},
  author={Demir, Ilke and Koperski, Krzysztof and Lindenbaum, David and Pang, Guan and Huang, Jing and Basu, Saikat and Hughes, Forest and Tuia, Devis and Raskar, Ramesh},
  booktitle={CVPR workshops},
  pages={172--181},
  year={2018}
}

@article{tong2023enabling,
  title={Enabling country-scale land cover mapping with meter-resolution satellite imagery},
  author={Tong, Xin-Yi and Xia, Gui-Song and Zhu, Xiao Xiang},
  journal={ISPRS Journal of Photogrammetry and Remote Sensing},
  volume={196},
  pages={178--196},
  year={2023},
  publisher={Elsevier}
}

@article{garioud2023flair,
  title={FLAIR: a country-scale land cover semantic segmentation dataset from multi-source optical imagery},
  author={Garioud, Anatol and Gonthier, Nicolas and Landrieu, Loic and De Wit, Apolline and Valette, Marion and Poup{\'e}e, Marc and Giordano, S{\'e}bastien and others},
  journal={NeurIPS},
  volume={36},
  pages={16456--16482},
  year={2023}
}

@ARTICLE{hypersigma,
  author={Wang, Di and Hu, Meiqi and Jin, Yao and Miao, Yuchun and Yang, Jiaqi and Xu, Yichu and Qin, Xiaolei and Ma, Jiaqi and Sun, Lingyu and Li, Chenxing and Fu, Chuan and Chen, Hongruixuan and Han, Chengxi and Yokoya, Naoto and Zhang, Jing and Xu, Minqiang and Liu, Lin and Zhang, Lefei and Wu, Chen and Du, Bo and Tao, Dacheng and Zhang, Liangpei},
  journal={IEEE Transactions on Pattern Analysis and Machine Intelligence}, 
  title={HyperSIGMA: Hyperspectral Intelligence Comprehension Foundation Model}, 
  year={2025},
  volume={47},
  number={8},
  pages={6427-6444}
}

@article{GID2020,
title = {Land-Cover Classification with High-Resolution Remote Sensing Images Using Transferable Deep Models},
author = {Xin-Yi Tong and Gui-Song Xia and Qikai Lu and Huangfeng Shen and Shengyang Li and Shucheng You and Liangpei Zhang},
journal = {Remote Sensing of Environment, doi: 10.1016/j.rse.2019.111322},
year = {2020}
}

@article{shi2023globe230k,
  title={Globe230k: A benchmark dense-pixel annotation dataset for global land cover mapping},
  author={Shi, Qian and He, Da and Liu, Zhengyu and Liu, Xiaoping and Xue, Jingqian},
  journal={Journal of Remote Sensing},
  volume={3},
  pages={0078},
  year={2023},
  publisher={AAAS}
}

@article{castillo2022semi,
  title={Semi-supervised semantic segmentation in earth observation: The minifrance suite, dataset analysis and multi-task network study},
  author={Castillo-Navarro, Javiera and Le Saux, Bertrand and Boulch, Alexandre and Audebert, Nicolas and Lef{\`e}vre, S{\'e}bastien},
  journal={Machine Learning},
  volume={111},
  number={9},
  pages={3125--3160},
  year={2022},
  publisher={Springer}
}

@article{sun2022fair1m,
  title={FAIR1M: A benchmark dataset for fine-grained object recognition in high-resolution remote sensing imagery},
  author={Sun, Xian and Wang, Peijin and Yan, Zhiyuan and Xu, Feng and Wang, Ruiping and Diao, Wenhui and Chen, Jin and Li, Jihao and Feng, Yingchao and Xu, Tao and others},
  journal={ISPRS Journal of Photogrammetry and Remote Sensing},
  volume={184},
  pages={116--130},
  year={2022},
  publisher={Elsevier}
}

@article{shu2025earthmind,
  title={EarthMind: Towards Multi-Granular and Multi-Sensor Earth Observation with Large Multimodal Models},
  author={Shu, Yan and Ren, Bin and Xiong, Zhitong and Paudel, Danda Pani and Van Gool, Luc and Demir, Begum and Sebe, Nicu and Rota, Paolo},
  journal={arXiv preprint arXiv:2506.01667},
  year={2025}
}

@inproceedings{quenumlisat,
  title={LISAt: Language-Instructed Segmentation Assistant for Satellite Imagery},
  author={Quenum, Jerome and Hsieh, Wen-Han and Wu, Tsung-Han and Gupta, Ritwik and Darrell, Trevor and Chan, David M},
  booktitle={The Thirty-ninth Annual Conference on Neural Information Processing Systems Datasets and Benchmarks Track}
}

@article{dong2024cross,
  title={Cross-modal bidirectional interaction model for referring remote sensing image segmentation},
  author={Dong, Zhe and Sun, Yuzhe and Liu, Tianzhu and Zuo, Wangmeng and Gu, Yanfeng},
  journal={arXiv preprint arXiv:2410.08613},
  year={2024}
}

@inproceedings{yu2016modeling,
  title={Modeling context in referring expressions},
  author={Yu, Licheng and Poirson, Patrick and Yang, Shan and Berg, Alexander C and Berg, Tamara L},
  booktitle={ECCV},
  pages={69--85},
  year={2016},
  organization={Springer}
}

@article{li2023phi15,
  title={Textbooks are all you need ii: phi-1.5 technical report},
  author={Li, Yuanzhi and Bubeck, S{\'e}bastien and Eldan, Ronen and Del Giorno, Allie and Gunasekar, Suriya and Lee, Yin Tat},
  journal={arXiv preprint arXiv:2309.05463},
  year={2023}
}

@inproceedings{liu2021swin,
  title={Swin transformer: Hierarchical vision transformer using shifted windows},
  author={Liu, Ze and Lin, Yutong and Cao, Yue and Hu, Han and Wei, Yixuan and Zhang, Zheng and Lin, Stephen and Guo, Baining},
  booktitle={ICCV},
  pages={10012--10022},
  year={2021}
}

@inproceedings{cheng2022masked,
  title={Masked-attention mask transformer for universal image segmentation},
  author={Cheng, Bowen and Misra, Ishan and Schwing, Alexander G and Kirillov, Alexander and Girdhar, Rohit},
  booktitle={CVPR},
  pages={1290--1299},
  year={2022}
}

@inproceedings{bengio2009curriculum,
  title={Curriculum learning},
  author={Bengio, Yoshua and Louradour, J{\'e}r{\^o}me and Collobert, Ronan and Weston, Jason},
  booktitle={ICML},
  pages={41--48},
  year={2009}
}

@inproceedings{ren2024pixellm,
  title={Pixellm: Pixel reasoning with large multimodal model},
  author={Ren, Zhongwei and Huang, Zhicheng and Wei, Yunchao and Zhao, Yao and Fu, Dongmei and Feng, Jiashi and Jin, Xiaojie},
  booktitle={CVPR},
  pages={26374--26383},
  year={2024}
}

@inproceedings{wanggeollava,
  title={GeoLLaVA-8K: Scaling Remote-Sensing Multimodal Large Language Models to 8K Resolution},
  author={Wang, Fengxiang and Chen, Mingshuo and Li, Yueying and Wang, Di and Wang, Haotian and Guo, Zonghao and Wang, Zefan and Boqi, Shan and Lan, Long and Wang, Yulin and others},
  booktitle={The Thirty-ninth Annual Conference on Neural Information Processing Systems}
}

@inproceedings{radford2021CLIP,
  title={Learning transferable visual models from natural language supervision},
  author={Radford, Alec and Kim, Jong Wook and Hallacy, Chris and Ramesh, Aditya and Goh, Gabriel and Agarwal, Sandhini and Sastry, Girish and Askell, Amanda and Mishkin, Pamela and Clark, Jack and others},
  booktitle={ICML},
  pages={8748--8763},
  year={2021},
  organization={PmLR}
}

@article{liu2024remoteclip,
  title={Remoteclip: A vision language foundation model for remote sensing},
  author={Liu, Fan and Chen, Delong and Guan, Zhangqingyun and Zhou, Xiaocong and Zhu, Jiale and Ye, Qiaolin and Fu, Liyong and Zhou, Jun},
  journal={IEEE Transactions on Geoscience and Remote Sensing},
  volume={62},
  pages={1--16},
  year={2024},
  publisher={IEEE}
}

@article{chiang2023vicuna,
  title={Vicuna: An open-source chatbot impressing gpt-4 with 90\%* chatgpt quality},
  author={Chiang, Wei-Lin and Li, Zhuohan and Lin, Ziqing and Sheng, Ying and Wu, Zhanghao and Zhang, Hao and Zheng, Lianmin and Zhuang, Siyuan and Zhuang, Yonghao and Gonzalez, Joseph E and others},
  journal={See https://vicuna. lmsys. org (accessed 14 April 2023)},
  volume={2},
  number={3},
  pages={6},
  year={2023}
}

@inproceedings{chen2024internvl,
  title={Internvl: Scaling up vision foundation models and aligning for generic visual-linguistic tasks},
  author={Chen, Zhe and Wu, Jiannan and Wang, Wenhai and Su, Weijie and Chen, Guo and Xing, Sen and Zhong, Muyan and Zhang, Qinglong and Zhu, Xizhou and Lu, Lewei and others},
  booktitle={CVPR},
  pages={24185--24198},
  year={2024}
}

@article{cai2024internlm2,
  title={Internlm2 technical report},
  author={Cai, Zheng and Cao, Maosong and Chen, Haojiong and Chen, Kai and Chen, Keyu and Chen, Xin and Chen, Xun and Chen, Zehui and Chen, Zhi and Chu, Pei and others},
  journal={arXiv preprint arXiv:2403.17297},
  year={2024}
}

@article{touvron2023llama,
  title={Llama 2: Open foundation and fine-tuned chat models},
  author={Touvron, Hugo and Martin, Louis and Stone, Kevin and Albert, Peter and Almahairi, Amjad and Babaei, Yasmine and Bashlykov, Nikolay and Batra, Soumya and Bhargava, Prajjwal and Bhosale, Shruti and others},
  journal={arXiv preprint arXiv:2307.09288},
  year={2023}
}

@inproceedings{ravi2024sam2,
  title={SAM 2: Segment Anything in Images and Videos},
  author={Ravi, Nikhila and Gabeur, Valentin and Hu, Yuan-Ting and Hu, Ronghang and Ryali, Chaitanya and Ma, Tengyu and Khedr, Haitham and R{\"a}dle, Roman and Rolland, Chloe and Gustafson, Laura and others},
  booktitle={ICLR},
  year={2024}
}

@inproceedings{sam_hq,
    title={Segment Anything in High Quality},
    author={Ke, Lei and Ye, Mingqiao and Danelljan, Martin and Liu, Yifan and Tai, Yu-Wing and Tang, Chi-Keung and Yu, Fisher},
    booktitle={NeurIPS},
    year={2023}
}

@ARTICLE{spex,
  author={Si, Dongchen and Wang, Di and Gao, Erzhong and Qin, Xiaolei and Zhao, Liu and Zhang, Jing and Xu, Minqiang and Zhan, Jianbo and Wang, Jianshe and Liu, Lin and Du, Bo and Zhang, Liangpei},
  journal={IEEE Transactions on Geoscience and Remote Sensing}, 
  title={SPEX: A Vision-Language Model for Land Cover Extraction on Spectral Remote Sensing Images}, 
  year={2026},
  volume={},
  number={},
  pages={1-1}
  }

@article{RSVG-ZeroOV,
  title={RSVG-ZeroOV: Exploring a Training-Free Framework for Zero-Shot Open-Vocabulary Visual Grounding in Remote Sensing Images},
  author={Li, Ke and Wang, Di and Wang, Ting and Dong, Fuyu and Zhang, Yiming and Zhang, Luyao and Wang, Xiangyu and Li, Shaofeng and Wang, Quan},
  journal={arXiv preprint arXiv:2509.18711},
  year={2025}
}

@article{DiffSegmenter,
  title={Diffusion model is secretly a training-free open vocabulary semantic segmenter},
  author={Wang, Jinglong and Li, Xiawei and Zhang, Jing and Xu, Qingyuan and Zhou, Qin and Yu, Qian and Sheng, Lu and Xu, Dong},
  journal={IEEE Transactions on Image Processing},
  year={2025},
  publisher={IEEE}
}

@inproceedings{DiffPNG,
  title={Exploring phrase-level grounding with text-to-image diffusion model},
  author={Yang, Danni and Dong, Ruohan and Ji, Jiayi and Ma, Yiwei and Wang, Haowei and Sun, Xiaoshuai and Ji, Rongrong},
  booktitle={ECCV},
  pages={161--180},
  year={2024},
  organization={Springer}
}

@article{wang2024OV-VG,
  title={OV-VG: A benchmark for open-vocabulary visual grounding},
  author={Wang, Chunlei and Feng, Wenquan and Li, Xiangtai and Cheng, Guangliang and Lyu, Shuchang and Liu, Binghao and Chen, Lijiang and Zhao, Qi},
  journal={Neurocomputing},
  volume={591},
  pages={127738},
  year={2024},
  publisher={Elsevier}
}

@inproceedings{shen2024groundvlp,
  title={Groundvlp: Harnessing zero-shot visual grounding from vision-language pre-training and open-vocabulary object detection},
  author={Shen, Haozhan and Zhao, Tiancheng and Zhu, Mingwei and Yin, Jianwei},
  booktitle={AAAI},
  volume={38},
  number={5},
  pages={4766--4775},
  year={2024}
}

@misc{zhou2025geogroundunifiedlargevisionlanguage,
      title={GeoGround: A Unified Large Vision-Language Model for Remote Sensing Visual Grounding}, 
      author={Yue Zhou and Mengcheng Lan and Xiang Li and Litong Feng and Yiping Ke and Xue Jiang and Qingyun Li and Xue Yang and Wayne Zhang},
      year={2025},
      eprint={2411.11904},
      archivePrefix={arXiv},
      primaryClass={cs.CV},
      url={https://arxiv.org/abs/2411.11904}, 
}

@inproceedings{zhang2023nextchatlmmchatdetection,
  title={NExT-Chat: An LMM for Chat, Detection and Segmentation},
  author={Zhang, Ao and Yao, Yuan and Ji, Wei and Liu, Zhiyuan and Chua, Tat-Seng},
  booktitle={ICML},
  pages={60116--60133},
  year={2024},
  organization={PMLR}
}

@article{bi2024deepseek,
  title={Deepseek llm: Scaling open-source language models with longtermism},
  author={Bi, Xiao and Chen, Deli and Chen, Guanting and Chen, Shanhuang and Dai, Damai and Deng, Chengqi and Ding, Honghui and Dong, Kai and Du, Qiushi and Fu, Zhe and others},
  journal={arXiv preprint arXiv:2401.02954},
  year={2024}
}

@article{wang2023hierarchical,
  title={Hierarchical open-vocabulary universal image segmentation},
  author={Wang, Xudong and Li, Shufan and Kallidromitis, Konstantinos and Kato, Yusuke and Kozuka, Kazuki and Darrell, Trevor},
  journal={NeurIPS},
  volume={36},
  pages={21429--21453},
  year={2023}
}

@ARTICLE{crossearth,
  author={Gong, Ziyang and Wei, Zhixiang and Wang, Di and Hu, Xiaoxing and Ma, Xianzheng and Chen, Hongruixuan and Jia, Yuru and Deng, Yupeng and Ji, Zhenming and Zhu, Xiangwei and Yang, Xue and Yokoya, Naoto and Zhang, Jing and Du, Bo and Yan, Junchi and Zhang, Liangpei},
  journal={IEEE Transactions on Pattern Analysis and Machine Intelligence}, 
  title={CrossEarth: Geospatial Vision Foundation Model for Domain Generalizable Remote Sensing Semantic Segmentation}, 
  year={2025},
  volume={},
  number={},
  pages={1-18}
  }

@article{S5,
  title={S5: Scalable Semi-Supervised Semantic Segmentation in Remote Sensing},
  author={Liang Lv and Di Wang and Jing Zhang and Lefei Zhang},
  journal={arXiv preprint arXiv:2508.12409},
  year={2025}
}

@inproceedings{wangsegllm,
  title={SegLLM: Multi-round Reasoning Segmentation with Large Language Models},
  author={Wang, XuDong and Zhang, Shaolun and Li, Shufan and Li, Kehan and Kallidromitis, Konstantinos and Kato, Yusuke and Kozuka, Kazuki and Darrell, Trevor},
  booktitle={ICLR}
}

@inproceedings{wang2025deep,
  title={Deep Adaptive Unfolded Network via Spatial Morphology Stripping and Spectral Filtration for Pan-sharpening},
  author={Wang, Hebaixu and Ma, Jiayi},
  booktitle={ICCV},
  pages={10730--10740},
  year={2025}
}

@inproceedings{wangdgsolver,
  title={DGSolver: Diffusion Generalist Solver with Universal Posterior Sampling for Image Restoration},
  author={Wang, Hebaixu and Zhang, Jing and Guo, Haonan and Wang, Di and Ma, Jiayi and Du, Bo},
  booktitle={The Thirty-ninth Annual Conference on Neural Information Processing Systems}
}

@article{wang2025residual,
  title={Residual Diffusion Bridge Model for Image Restoration},
  author={Wang, Hebaixu and Zhang, Jing and Chen, Haoyang and Guo, Haonan and Wang, Di and Ma, Jiayi and Du, Bo},
  journal={arXiv preprint arXiv:2510.23116},
  year={2025}
}
}

\clearpage
\setcounter{page}{1}
\maketitlesupplementary

\section{Overview}
This supplementary material provides further details for the proposed dataset GeoSeg-1M and benchmark GeoSeg-Bench, as well as the proposed model UniGeoSeg. These details were omitted from the main body of the paper due to space constraints.

The supplementary material is organized as follows:

\begin{itemize}
\item Section~\ref{sec:Details of the GeoSeg-1M Dataset Creation.}. Details of the GeoSeg-1M Dataset Creation.

\item Section~\ref{sec:The Statistics of GeoSeg-1M}. Statistics and Additional Samples of GeoSeg-1M.

\item Section~\ref{sec:The statistics of GeoSeg-Bench}. More Details about GeoSeg-Bench.

\item Section~\ref{sec:More Details about Experiments}. More Details about Experiments.

\item Section~\ref{sec:Additional Examples of Model Predictions}. Additional Examples of Model Predictions.

\item Section~\ref{sec:Evaluating UniGeoSeg with Alternative Language Model}. Evaluating UniGeoSeg with Alternative Language Model.

\item Section~\ref{app-datasheets} Datasheets for GeoSeg-1M and GeoSeg-Bench.

\end{itemize}

\section{Details of the GeoSeg-1M Dataset Creation}
\label{sec:Details of the GeoSeg-1M Dataset Creation.}
\subsection{Data Sources of GeoSeg-1M}
\label{sec:Data Sources of GeoSeg-1M}
To construct GeoSeg-1M, we aggregate a wide range of publicly available remote-sensing datasets that provide pixel-level or region-level annotations across diverse spatial resolutions, geographic regions, and semantic categories. These datasets collectively supply the raw images and masks from which our unified multimodal corpus is built. Below, we summarize the data sources used in GeoSeg-1M and provide brief descriptions of each dataset.

\textbf{Chesapeake Land Cover}~\cite{robinson2019large} provides a high-resolution (1 m) land-cover classification for the Chesapeake Bay watershed, covering regions in Maryland, Virginia, West Virginia, Pennsylvania, Delaware, and Washington D.C. The dataset includes raster layers derived from NAIP (Red, Green, Blue, NIR) imagery as well as Landsat-8 bands, with land-cover labels for classes such as water, tree canopy / forest, low vegetation, barren land, other impervious surfaces, roads, and no-data. The data was produced in collaboration between the Chesapeake Conservancy and USGS, and reflects coverage for multiple epochs (2013/2014, 2017/2018, and 2021/2022) with detailed land-use and land-cover change products.

\textbf{DeepGlobe}~\cite{demir2018deepglobe} dataset originates from the DeepGlobe 2018 Satellite Image Understanding Challenge and includes a land-cover segmentation track. For the segmentation / land-cover component, it provides 803 RGB satellite images of size 2448 × 2448 pixels with a ground sampling distance of approximately 0.5 m. The annotation consists of seven land-cover classes, including urban, agriculture, rangeland, forest, water, barren, and unknown. 

\textbf{FLAIR}~\cite{garioud2023flair} is a country-scale land-cover semantic segmentation dataset for France. It is built from multi-source optical imagery with 20 cm ground sampling distance, and contains more than 20 billion labeled pixels across more than 817 km² of aerial acquisitions. In addition to spatial data, FLAIR integrates temporal and spectral information from satellite time series, providing fine-grained annotations for land-use monitoring and segmentation research.

\textbf{GID-15}~\cite{GID2020} is a large-scale semantic segmentation dataset based on Gaofen-2 satellite images. It consists of 150 GF-2 images with pixel-level annotations covering 15 land-cover classes. The dataset provides two types of ground-truth format: .png (grayscale labels) and .tiff (RGB palette), to facilitate different usage scenarios.

\textbf{Globe230K}~\cite{shi2023globe230k} is a globally distributed land-cover dataset consisting of approximately 232,819 image tiles of size 512×512 pixels. The tiles have a 1-m ground sampling distance and are annotated with a set of land-use/land-cover categories covering diverse geographic regions.

\textbf{LoveDA}~\cite{wang2loveda} is a semantic segmentation dataset constructed from 0.3 m spatial-resolution images collected via the Google Earth platform. It covers both urban and rural regions, providing scenes of varied spatial layouts. The dataset includes six foreground classes, and offers pixel-level annotations for all categories.

\textbf{MiniFrance}~\cite{castillo2022semi} is a very-high-resolution aerial semantic segmentation dataset released for the IEEE Data Fusion Contest 2022 (DFC2022). It contains approximately 2,000 VHR images annotated with 12 land-use/land-cover classes, based on the Urban Atlas project. The training split of MiniFrance includes both labeled and unlabeled images to support semi-supervised learning.

\textbf{Potsdam}~\cite{bayanlou2021multi} is a benchmark aerial segmentation dataset consisting of 38 orthorectified tiles, each of size 6000×6000 pixels, acquired with a ground sampling distance of 5 cm. The images include four spectral bands (R, G, B, NIR), and the corresponding ground-truth labels cover six semantic categories: impervious surfaces, buildings, low vegetation, trees, cars, and background/clutter.

\textbf{Vaihingen}~\cite{bayanlou2021multi} is an aerial semantic segmentation dataset composed of 33 image patches, captured at a ground sampling distance of 9 cm. The dataset provides both true orthophotos and corresponding digital surface models, and its annotated classes include six categories: impervious surfaces, buildings, low vegetation, trees, cars, and clutter/background.

\textbf{Five-Billion-Pixels}~\cite{tong2023enabling} is a land-cover classification dataset derived from Gaofen-2 satellite imagery with a spatial resolution of 4 m. It contains more than five billion labeled pixels distributed across 150 large-area scenes and is annotated with 24 land-cover classes.

\textbf{FAIR1M}~\cite{sun2022fair1m} is a large-scale fine-grained object detection dataset in high-resolution remote sensing imagery. It contains over 15,000 images with a spatial resolution between 0.3 m and 0.8 m, collected from Gaofen-2 satellites and Google Earth. The dataset comprises more than 1 million object instances, annotated with rotated bounding boxes across 5 main categories (airplanes, ships, vehicles, courts, roads) and 37 sub-categories.

\textbf{DIOR}~\cite{li2020object} is a large-scale object detection dataset comprising 23,463 remote-sensing images annotated with 192,472 object instances from 20 categories. The images exhibit variations in viewpoints, scenes, and spatial resolutions, and each object is annotated using axis-aligned bounding boxes.

\textbf{DOTA}~\cite{xia2018dota} is an aerial image dataset for object detection containing 2,806 images with sizes ranging from 800×800 to 4000×4000 pixels. It includes 188,282 annotated object instances from 15 categories, where each object is labeled with a bounding box allowing arbitrary orientations.

For FAIR1M, DIOR, and DOTA, which provide region-level annotations in the form of bounding boxes rather than pixel-level masks, we convert the box annotations into segmentation masks using the procedure of SAMRS\cite{wang2023samrs}. This conversion allows all region-level annotations from FAIR1M, DIOR, and DOTA to be integrated into our unified pixel-based segmentation framework.

\textbf{RRSIS-D}~\cite{liu2024rotated} consists of 17,402 image–text–mask triplets intended for the referring segmentation task in remote sensing. The images in RRSIS-D have a size of 800×800 pixels, covering a variety of ground sampling distances (from ~0.5 m to 0.3 m), and the dataset includes 20 object categories described in natural-language expressions.

\textbf{EarthReason}~\cite{li2025segearthr1} is a benchmark for a novel remote-sensing task called geospatial pixel reasoning. The dataset contains 5,434 manually annotated image masks and over 30,000 implicit question-answer pairs, where each question describes a target region in an image in a reasoning style rather than direct naming. The dataset supports implicit querying, requiring models to infer masks from contextual and spatial clues rather than explicit instructions.

\textbf{RemoteSAM-270K}~\cite{yao2025remotesam} is a large-scale generalized referring expression segmentation dataset for remote sensing, consisting of 270K image-text-mask triplets. It covers 297 object categories and 16 fine-grained attribute types with an average of 3.17 attributes per sample, supporting multi-target, no-target, and single-target expressions.

\subsection{Data Pre-processing of GeoSeg-1M}
\label{sec:Data Pre-processing of GeoSeg-1M}
The raw data used to construct GeoSeg-1M originate from multiple remote sensing benchmarks with substantial variations in spatial resolution, sensor characteristics, annotation formats, and category taxonomies. These datasets typically provide large aerial images paired with semantic segmentation masks encoded in heterogeneous ways (\eg, indexed masks, palette masks, or RGB-encoded labels). Such inconsistencies make unified processing and subsequent multimodal instruction generation difficult. To ensure compatibility across sources and create a standardized corpus suitable for multimodal segmentation, we perform a unified preprocessing pipeline on all images and masks.

\begin{figure*}[h]
    \centering
    \includegraphics[width=1.0\linewidth]{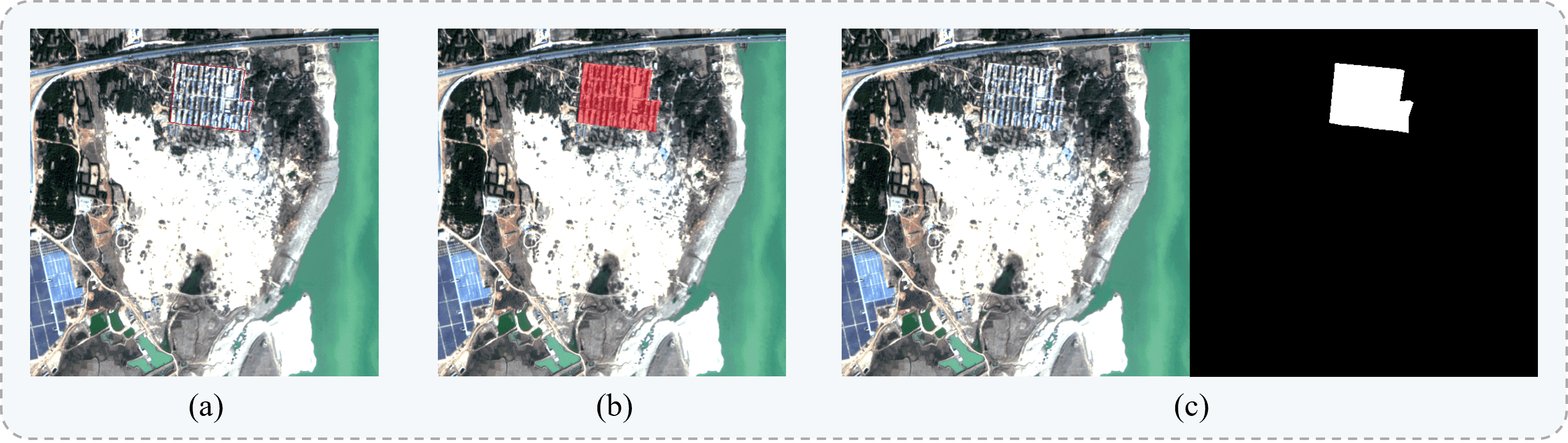}
    \caption{The three mask marking strategies we tried in model-based mask filtering. (a) Boundary-only highlight. (b) Semi-transparent filled-mask overlay. (c) The original image and the binary mask}
    \label{fig:supp_mask_filter_threemethod}
\end{figure*}

\begin{figure*}[h]
    \centering
    \includegraphics[width=1.0\linewidth]{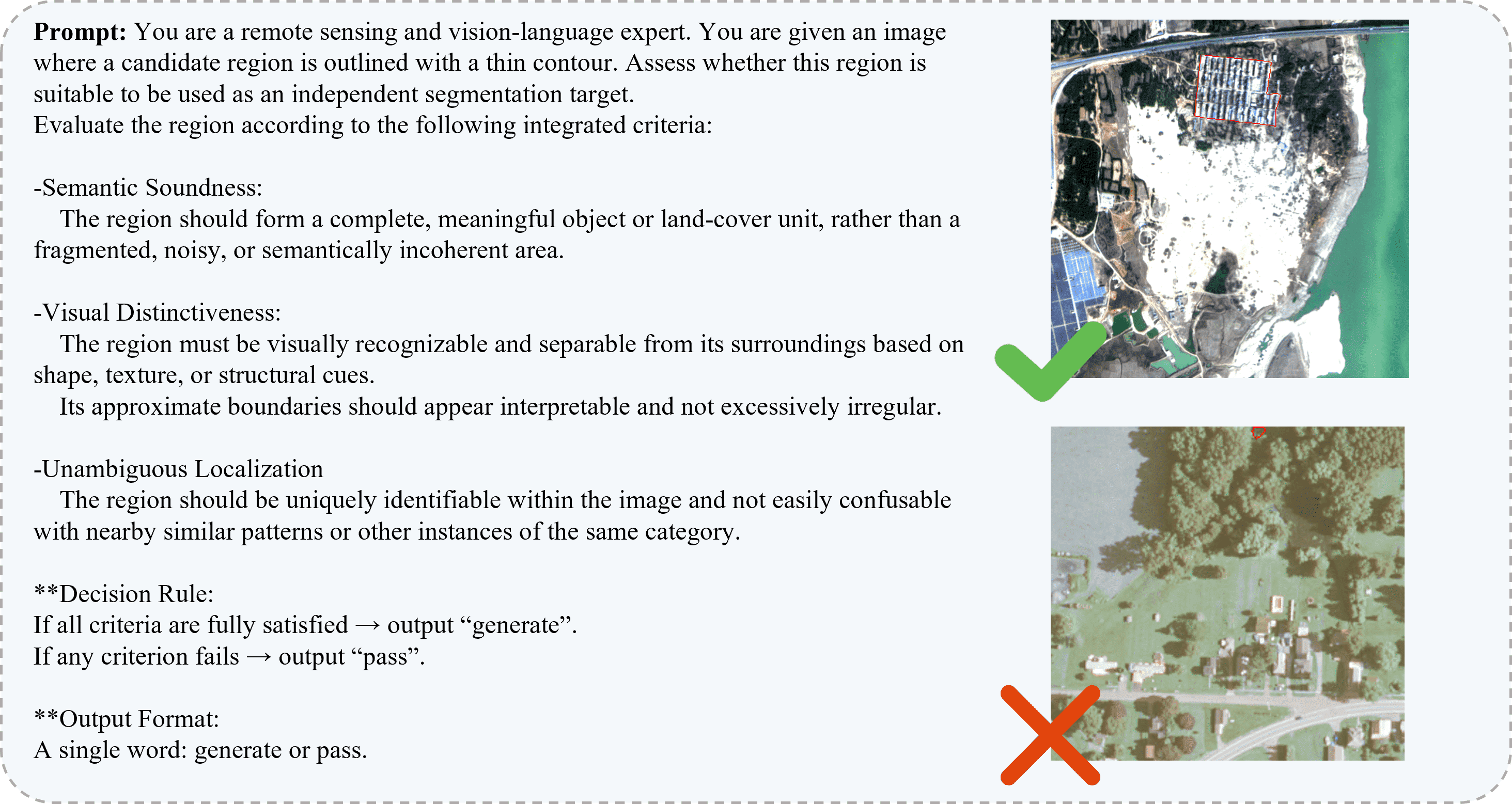}
    \caption{The prompt of InternVL3 for mask filtering. The images marked with $\checkmark$ in the top-right corner are retained, as their mask regions exhibit clear semantic structure, while the images marked with $\times$ in the bottom-right corner are filtered out due to ambiguous semantics.}
    \label{fig:supp_mask_filtering}
\end{figure*}

\textbf{Patch Extraction.} We adopt a sliding-window strategy to extract fixed-size patches from both images and masks. Each image is divided into patches of 512 × 512 pixels with a stride of 256. The corresponding masks are cropped using identical window parameters to maintain spatial alignment.

\textbf{Mask Standardization.} All annotations are converted into single-channel categorical masks with integer IDs. Regardless of the original encoding format, every pixel is mapped to a unified category index. This enables uniform downstream parsing while preserving the semantics of each dataset. We then consolidate all category label definitions into a unified taxonomy. Through synonym merging, we produce a standardized set of 117 semantic categories. This harmonization ensures that patches originating from different datasets share a consistent semantic space and can be used jointly for training a single segmentation model.

\textbf{Instance Candidate Extraction.} To support object-centric instruction-driven tasks such as referring segmentation, we further decompose the semantic masks into instance-level candidates. Since most remote sensing datasets provide class-wise semantic masks rather than instance annotations, we apply connected-component analysis (using 8-connectivity) to each semantic category map. Each connected region is treated as a potential instance for subsequent evaluation and instruction generation. Only minimal sanity checks—such as removing extremely tiny isolated regions or invalid labels—are applied at this stage; the full quality assessment is deferred to the mask filtering pipeline described in the next subsection.

This preprocessing stage yields a unified collection of standardized image patches, normalized masks, harmonized categories, and instance-level region proposals, which together form the basis for the subsequent high-quality mask filtering and instruction construction procedures.

\begin{figure*}[h]
    \centering
    \includegraphics[width=1.0\linewidth]{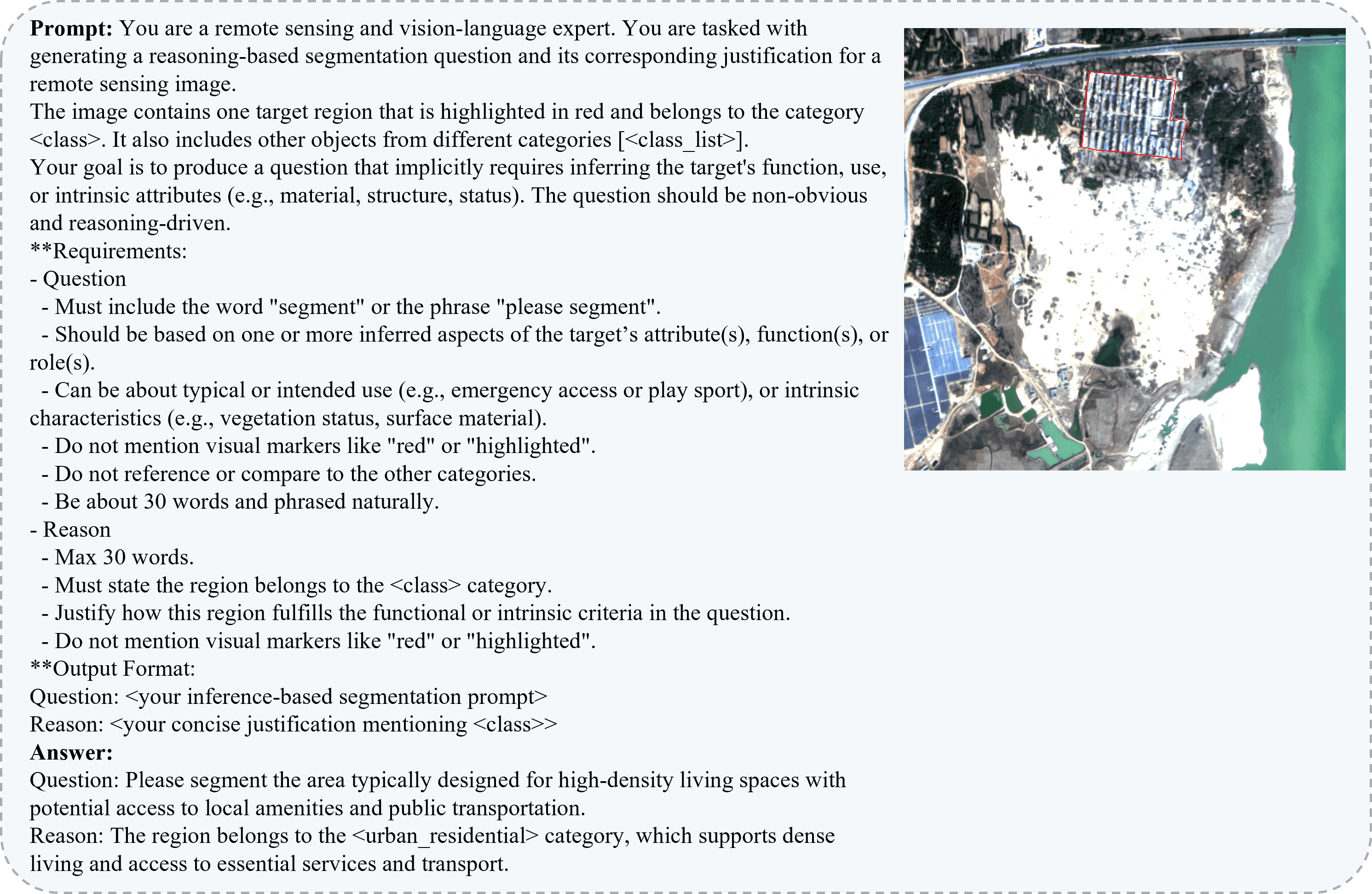}
    \caption{The prompt of GPT-4o for attribute reasoning instruction generation.}
    \label{fig:supp_attr_gen}
\end{figure*}

\subsection{Mask Filtering of GeoSeg-1M}
\label{sec:Mask Filtering of GeoSeg-1M}
After obtaining instance-level region proposals from connected-component decomposition, we perform a two-stage mask filtering pipeline to ensure that only high-quality, semantically meaningful regions are preserved for instruction generation. This process is designed to remove noisy, ambiguous, or low-value regions that commonly arise from fragmented masks and inconsistent labeling in remote sensing datasets.

\textbf{Rule-Based Filtering.} In the first stage, we apply rule-based filtering guided by geometric and contextual constraints. Each instance mask must occupy between 0.5\% and 70\% of the patch area; extremely small regions provide insufficient visual cues while excessively large ones often cover entire landcover zones that are unsuitable for instruction-driven segmentation tasks. To avoid ambiguity in referring and reasoning tasks, instances from the same category are restricted as follows:

\noindent(i) Each region must not belong to a category with more than six instances in the same patch, preventing confusion when many small, similar objects cluster together. 

\noindent(ii) For each candidate region, no other instance of the same category may lie within a 15-pixel radius, ensuring that fragmented or closely adjacent subregions are not incorrectly treated as distinct objects.

\noindent(iii) To maintain dataset diversity and prevent over-representation of dominant classes, at most two instances per category in each patch are randomly sampled to enter the next evaluation stage. These constraints effectively suppress low-quality regions while preserving category variety across the dataset.

\textbf{Model-Based Filtering.} In the second stage, we perform a model-based assessment using InternVL3 to further evaluate whether each region can serve as a meaningful standalone segmentation target. Rather than feeding raw binary masks—which completely fill the region and obscure internal appearance—we generate an outline-based visualization that draws a thin contour around the candidate region while preserving all underlying image content. This representation keeps both the object and its surroundings fully visible, providing richer cues for quality assessment.

Before settling on this approach, we tried three mask marking strategies: (i) boundary-only highlighting, (ii) semi-transparent filled-mask overlays, and (iii) supplying the original image and the mask as separate inputs, as shown in Fig.~\ref{fig:supp_mask_filter_threemethod}. Human inspection and initial model responses indicated that boundary-only visualization offers the most reliable evaluations. Filled overlays tend to partially hide texture and boundary details, while separate-mask inputs reduce contextual awareness. In contrast, the outline-based representation maintains precise localization cues without compromising scene context. Using this visualization, InternVL3 is prompted to judge semantic soundness, visual distinctiveness, and unambiguous localization of the region, based on the instruction shown in Fig.~\ref{fig:supp_mask_filtering}. Only regions validated by the model proceed to instruction construction.

Together, this two-stage filtering pipeline substantially improves annotation quality by removing ambiguous, noisy, or low-information regions and ensuring that the retained masks correspond to visually clear, contextually meaningful, and well-defined objects suitable for instruction-driven segmentation tasks.

\begin{figure*}[h]
    \centering
    \includegraphics[width=1.0\linewidth]{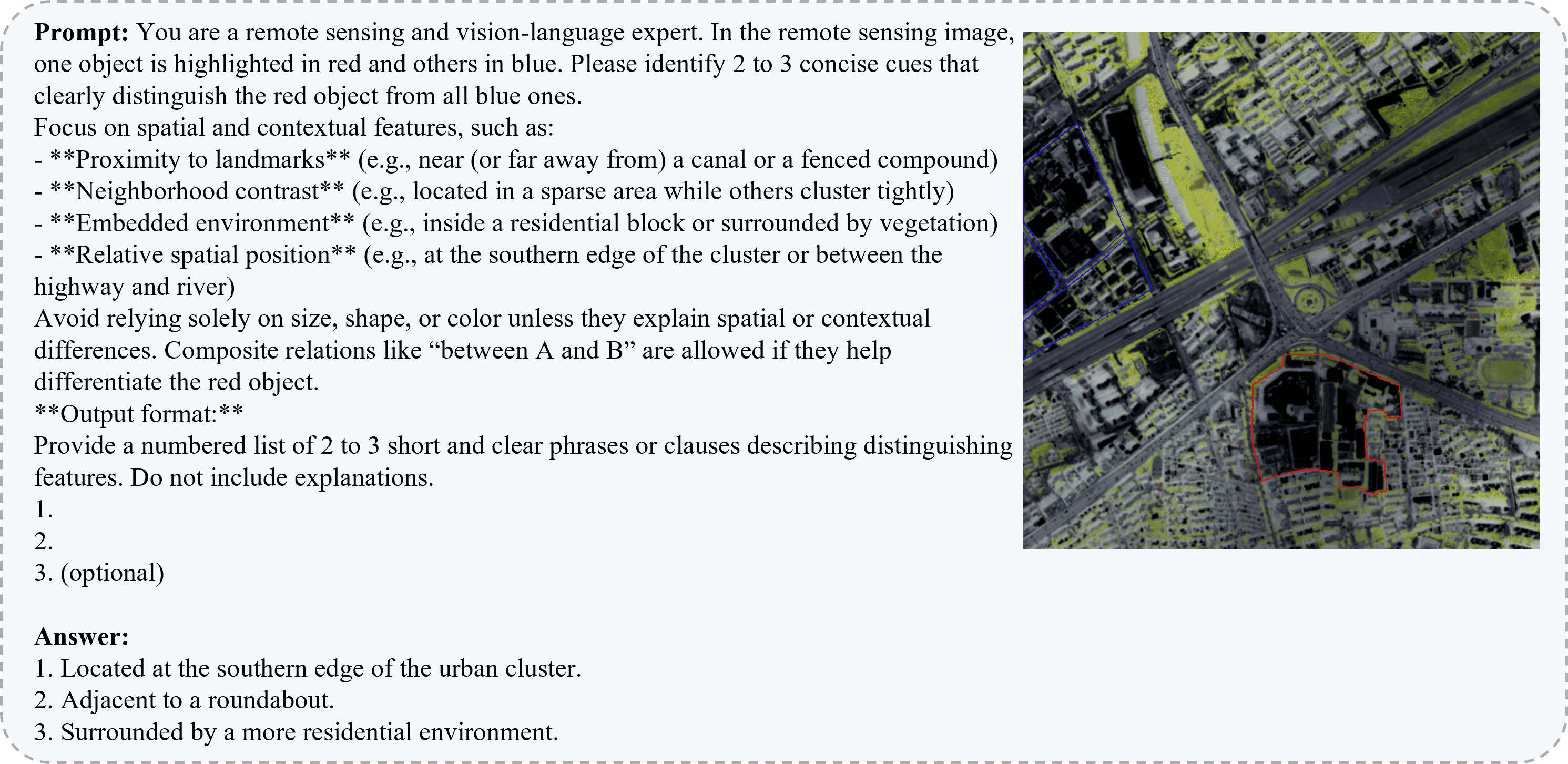}
    \caption{The prompt of GPT-4o for the first step of context reasoning instruction generation.}
    \label{fig:supp_context_step1}
\end{figure*}

\begin{figure*}[h]
    \centering
    \includegraphics[width=1.0\linewidth]{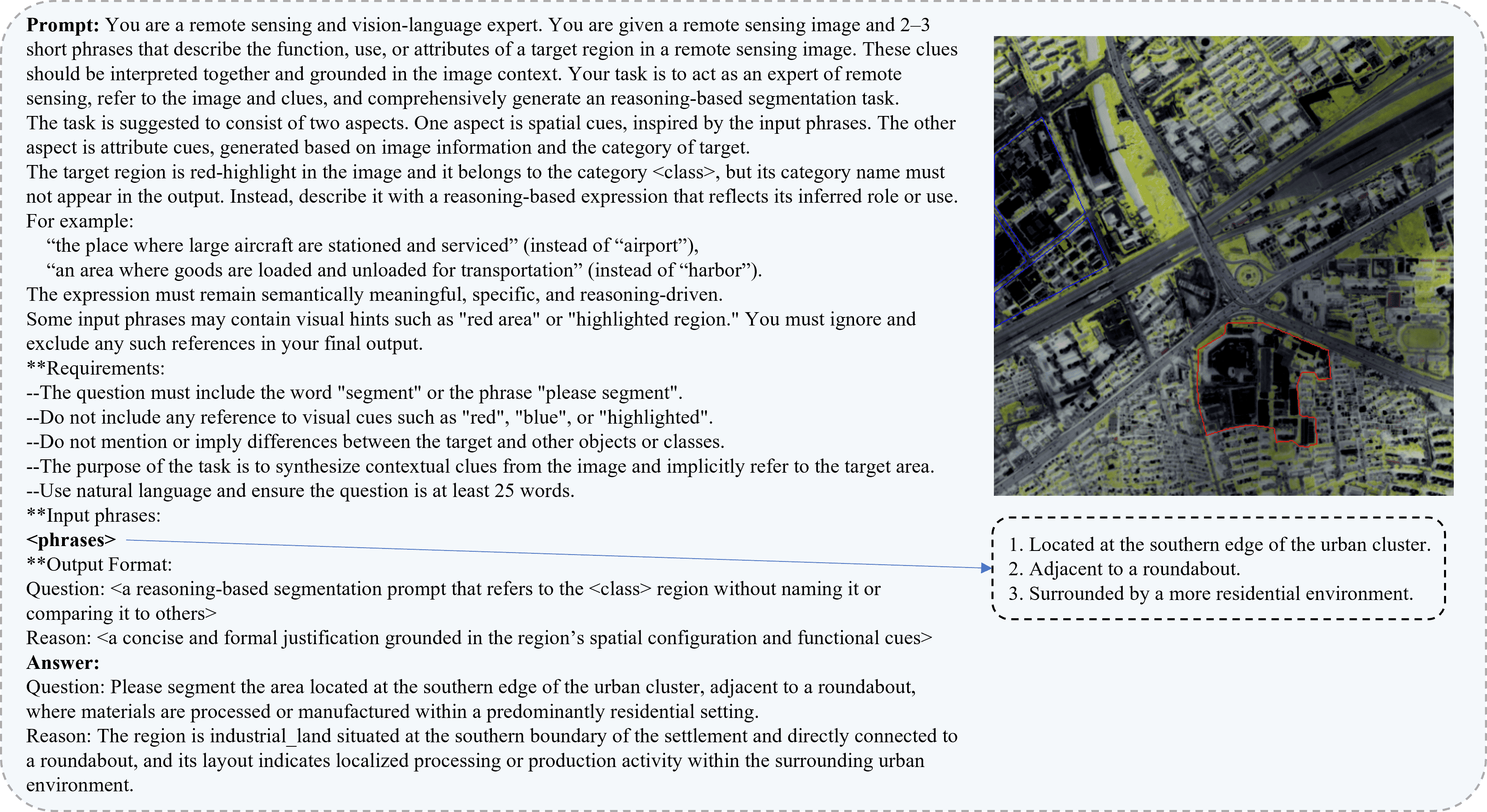}
    \caption{The prompt of GPT-4o for the second step of context reasoning instruction generation.}
    \label{fig:supp_context_step2}
\end{figure*}

\begin{figure*}[ht]
    \centering
    \includegraphics[width=1.0\linewidth]{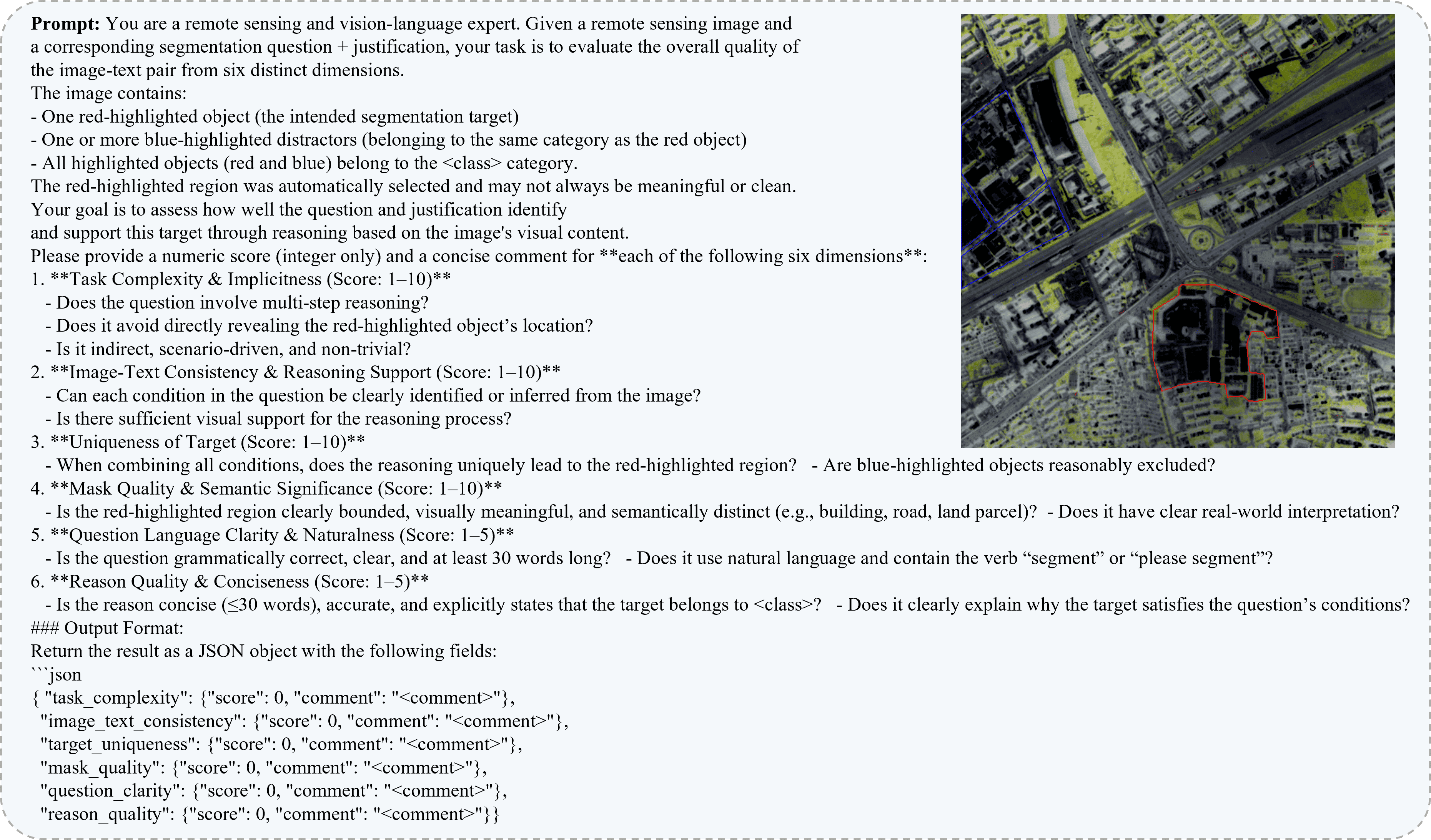}
    \caption{The prompt of InternVL3 and QwenVL2 for evaluating the quality of reasoning image–mask–instruction triplets.}
    \label{fig:supp_score}
\end{figure*}

\subsection{Reasoning Instruction Generation}
\label{sec:Reasoning Instruction Generation}
To construct high-quality reasoning instructions, we design a multi-stage pipeline that selects category-diverse images, differentiates attribute reasoning from contextual reasoning cases, and employs both GPT-4o and open-source VLMs for generation and cross-evaluation, respectively.

\textbf{Semantic Diversity Filtering.} We first ensure that the images entering this stage possess sufficiently diverse semantic content. For each dataset, we compute the empirical distribution of object categories based on the corresponding masks and rank images by the number of distinct categories they contain. Only images within the high-diversity tier of the distribution of each dataset are preserved for further processing. Although exact thresholds differ due to dataset-specific statistics, they generally correspond to selecting roughly the top decile.

\textbf{Reasoning Type Assignment.} For every selected image, we determine the reasoning type by examining how many regions belong to same semantic category. Images containing a single instance of the target category are treated as attribute reasoning cases, whereas images containing two or three instances are designated as contextual reasoning cases. This separation enables us to distinguish samples that require reasoning about intrinsic properties from those involving relational or discriminative understanding. 

\textbf{Attribute Reasoning Prompt.} For attribute reasoning samples, GPT-4o receives two inputs—the image with the target region outlined using a thin contour and the list of all categories present in the image—and is instructed to produce a question centered on attributes, functional roles, or other intrinsic characteristics. The corresponding prompt template is shown in Fig.~\ref{fig:supp_attr_gen}.

\textbf{Context Reasoning Prompt.} Context reasoning samples involve a two-stage generation process. All same-category regions are highlighted using distinct colors to clearly separate them. GPT-4o first produces two to three explicit visual cues that distinguish the highlighted regions in terms of spatial arrangement, geometry, or appearance (prompt in Fig.~\ref{fig:supp_context_step1}). These cues, together with the original image, are then used in a second GPT-4o step, where the model generates a reasoning question and a concise explanation grounded in contextual or relational understanding (prompt in Fig.~\ref{fig:supp_context_step2}).

\begin{figure*}[ht]
    \centering
    \includegraphics[width=1.0\linewidth]{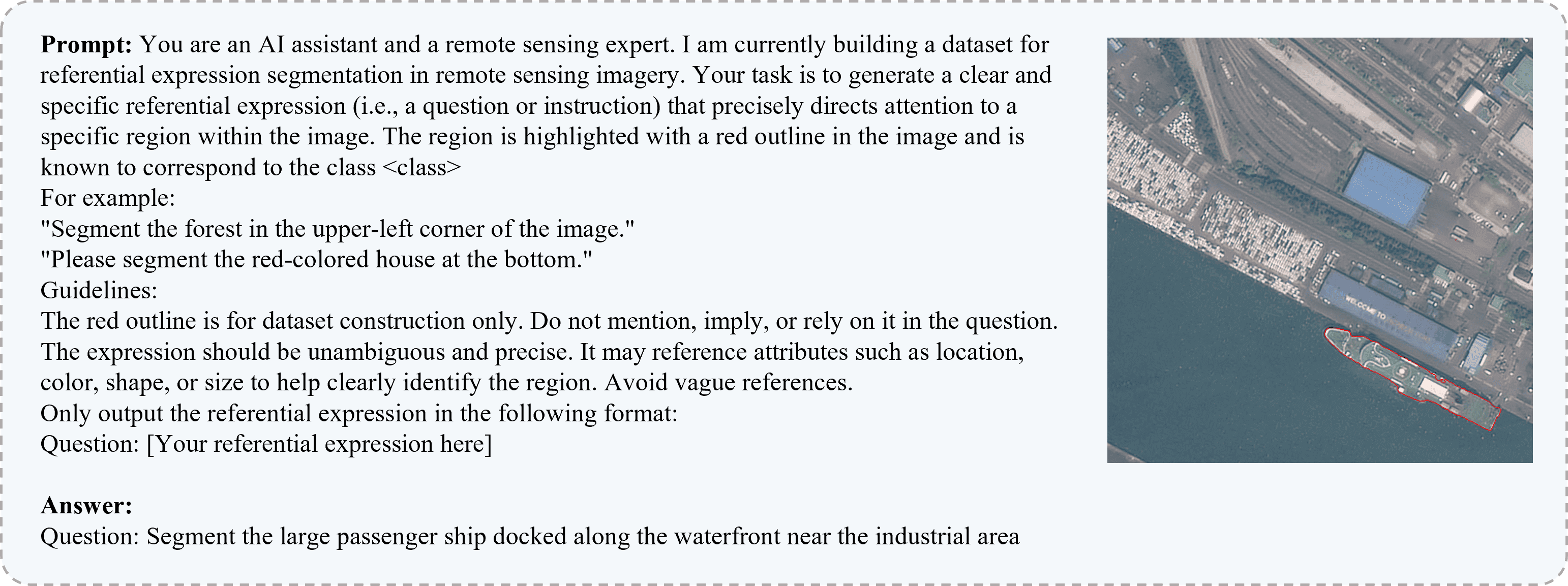}
    \caption{The prompt of GPT-4o for referring instruction generation.}
    \label{fig:supp_ref}
\end{figure*}

\begin{figure*}[ht]
    \centering
    \includegraphics[width=1.0\linewidth]{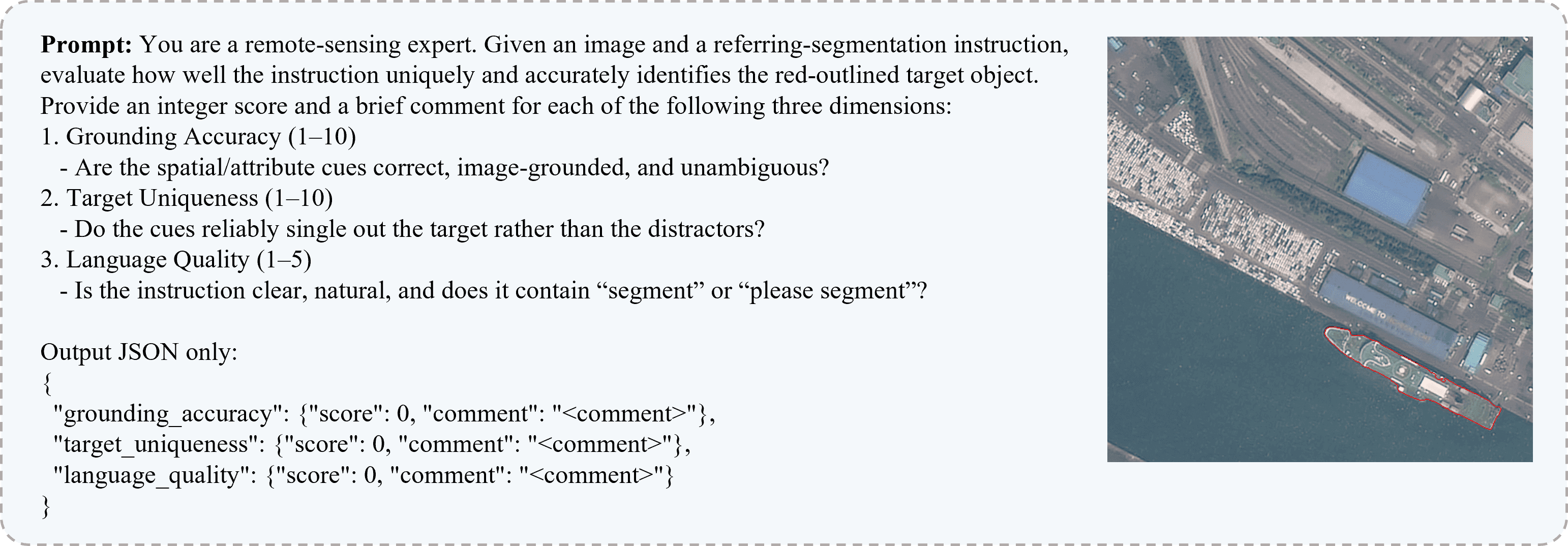}
    \caption{The prompt of InternVL3 and QwenVL2 for evaluating the quality of referring image–mask–instruction triplets.}
    \label{fig:supp_ref_score}
\end{figure*}

\begin{table*}[htbp]
\centering
\small
\caption{Statistics of GeoSeg-1M.}
\label{tab:geoseg_statistics}
\resizebox{\textwidth}{!}{%
\begin{tabular}{lccccccccc}
\toprule
\textbf{Subset}  & \textbf{Samples (generated)} & \textbf{Samples (total)} & \textbf{Avg Text Length} & \textbf{Categories} & \textbf{Average Mask Size} & \textbf{Other Statistics} \\
\midrule
Referring  & 336,311 & 560,179 & 12.05 words & 100 & 18.4 K  & - \\
Interactive  & 480,949 & 480,949 & 9.80 words & 106 & 15.5 K  & \makecell[c]{Point:242,488 \\ Box: 238,461} \\
Reasoning & 104,989 & 119,214 & 23.93 words & 108 & 14.8 K  & \makecell[c]{Attribute: 63,567 \\ Contextual: 41,422} \\
\midrule
Overall & 922,249 & 1,148,504 & 12.18 words & 117 & 17.0 K  & \makecell[c]{Resolution: 0.05--153 m \\ Sources$^1$: C, DG, FB, FL, GI, GL, L \\ M, P, Va, FA, DO, DI, EarthReason~\cite{li2025segearthr1}\\ RemoteSAM~\cite{yao2025remotesam}, RRSIS-D~\cite{liu2024rotated}} \\
\bottomrule
\end{tabular}%
}
\\[6pt] 
  \raggedright
\small{$1$ C:Chesapeake~\cite{robinson2019large}, DG:DeepGlobe~\cite{demir2018deepglobe}, FB:Five-Billion-Pixel~\cite{tong2023enabling}, FL:FLAIR~\cite{garioud2023flair}, GI:GID-15~\cite{GID2020}, GL:Globe230K~\cite{shi2023globe230k}, L:LoveDA~\cite{wang2loveda}, M:MiniFrance~\cite{castillo2022semi}, P:Potsdam \cite{bayanlou2021multi}, Va:Vaihingen \cite{bayanlou2021multi}, FA:FAIR1M \cite{sun2022fair1m}, DO:DOTA \cite{xia2018dota}, DI:DIOR \cite{li2020object}.}\\

\end{table*}

\textbf{Quality Filtering.} All resulting triplets of image–mask–instruction are further validated through cross-model evaluation using QwenVL2-72B and InternVL3-78B. Both models score each sample along multiple dimensions, including task complexity and implicitness, image–text consistency and reasoning soundness, uniqueness of the target, mask quality and semantic significance, clarity and naturalness of the question, and conciseness of the explanation. The scoring prompt is provided in Fig.~\ref{fig:supp_score}. Each dimension has a task-specific threshold, and only samples that exceed all thresholds across both evaluators are retained. After this filtering, the final reasoning subset contains 104,989 samples, including 63,567 attribute reasoning samples and 41,422 context reasoning samples.

\subsection{Referring Instruction Generation}
\label{sec:Referring Instruction Generation}

To construct the referring segmentation subset, we employ a unified prompting template that guides GPT-4o to produce concise and unambiguous referential expressions for each candidate region. As shown in Fig.~\ref{fig:supp_ref}, the prompt provides the image patch, the semantic class of the target region, and a strict output format beginning with “Question:”. The only emphasized requirement in the template is that the expression must clearly and uniquely identify the region based on interpretable cues.

A notable effect of this design is that the generated instructions naturally rely on contextual and spatial information—such as relative position, surrounding structures, or functional cues—rather than superficial attributes like color or size. This leads to more meaningful and semantically grounded referring expressions, consistent with real-world remote sensing interpretation.

Following instruction generation, all samples undergo cross-model validation using InternVL3-78B and QwenVL2-72B. Both models independently evaluate each instruction–image pair using the scoring prompt shown in Fig.~\ref{fig:supp_ref_score}, assessing clarity, grounding, consistency, and uniqueness of the reference. Only samples that satisfy the required thresholds across both evaluators are retained. After this stage, the final referring subset contains 336,311 high-quality region–instruction pairs.

\subsection{Interactive Instruction Generation}
\label{sec:Interactive Instruction Generation}
For interactive segmentation, we generate point- and box-based instructions from mask annotations. For each mask region, we sample \(k \in \{1,2,3\}\) points and normalize their coordinates to the \([0,1]\) range along both image axes. The number of points \(k\) is determined adaptively according to the mask size: regions with fewer than 200 pixels are assigned a single point, whereas larger regions follow a distribution in which single-point prompts account for 60\% of cases, and two- or three-point prompts each account for 20\%. Denoting the normalized coordinates of the sampled points as \((x_i, y_i)\), \(i=1,\dots,k\), the points are inserted into a fixed template:
\textit{``Please segment the region/target corresponding to the points \{($x_1$, $y_1$), \dots, ($x_k$, $y_k$)\}.''}

For each mask, we compute the tight bounding box defined by the top-left corner \((x_\text{min}, y_\text{min})\) and bottom-right corner \((x_\text{max}, y_\text{max})\)), and normalize all coordinates to \([0,1]\). These coordinates are then inserted into the template:
\textit{``Please segment the region/target corresponding to the box x0,y0=[$x_{min}$,$y_{min}$], x1,y1=[$x_{max}$,$y_{max}$].''}

This procedure produces 480,949 interactive samples, providing consistent and structured supervision for both point- and box-driven segmentation tasks.

\section{Statistics and Additional Samples of GeoSeg-1M}
\label{sec:The Statistics of GeoSeg-1M}

\begin{figure*}[th]
    \centering
    \includegraphics[width=\linewidth]{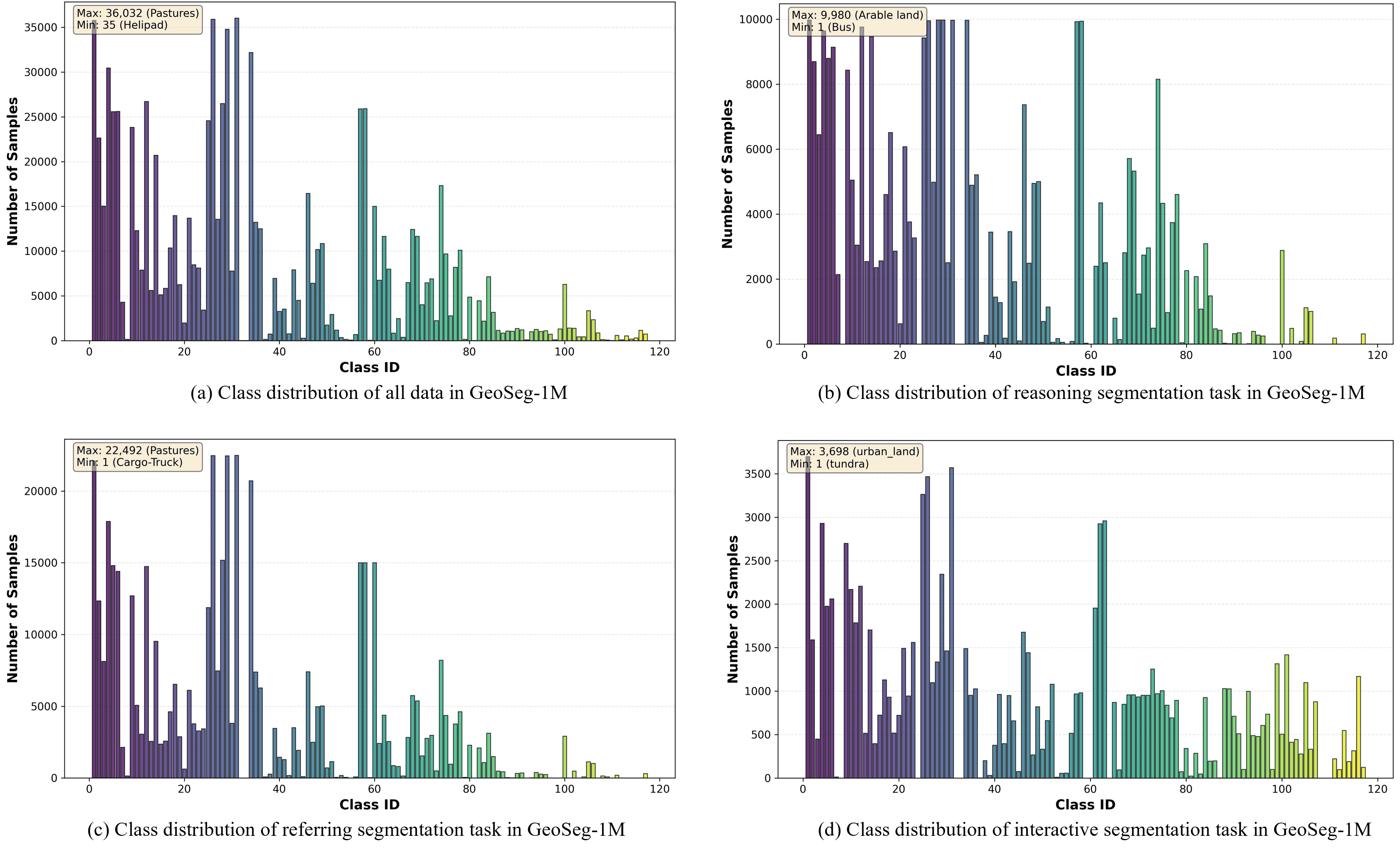}
    \caption{Class distribution in GeoSeg-1M. (a) Overall class distribution across the entire dataset. (b–d) Class distributions for each specific task: (b) Reasoning segmentation, (c) Referring segmentation, and (d) Interactive segmentation.}
    \label{fig:class_distribution_single}
\end{figure*}

Tab.~\ref{tab:geoseg_statistics} presents detailed statistics of the GeoSeg-1M dataset. The dataset is divided into three subsets corresponding to the main segmentation tasks: referring, interactive, and reasoning. For each subset, we report the number of samples generated in our pipeline, as well as the total number of samples including integrated external datasets. The average text length indicates the typical instruction length for each task, while the number of categories reflects the semantic diversity covered. We also provide the average mask size per subset and task-specific statistics, such as point/box distribution for interactive samples and attribute/context counts for reasoning samples. Overall, GeoSeg-1M contains more than 1.14 million image–mask–instruction triplets spanning 117 categories, aggregated from multiple remote sensing sources. The semantic categories are listed in Tab.~\ref{tab:geoseg_categories}, and their overall quantity distribution is shown in Fig.~\ref{fig:class_distribution_single}. The distributions of categories for the reasoning, referring, and interactive tasks are also presented separately.

\begin{figure*}[th]
    \centering
    \includegraphics[width=0.88\linewidth]{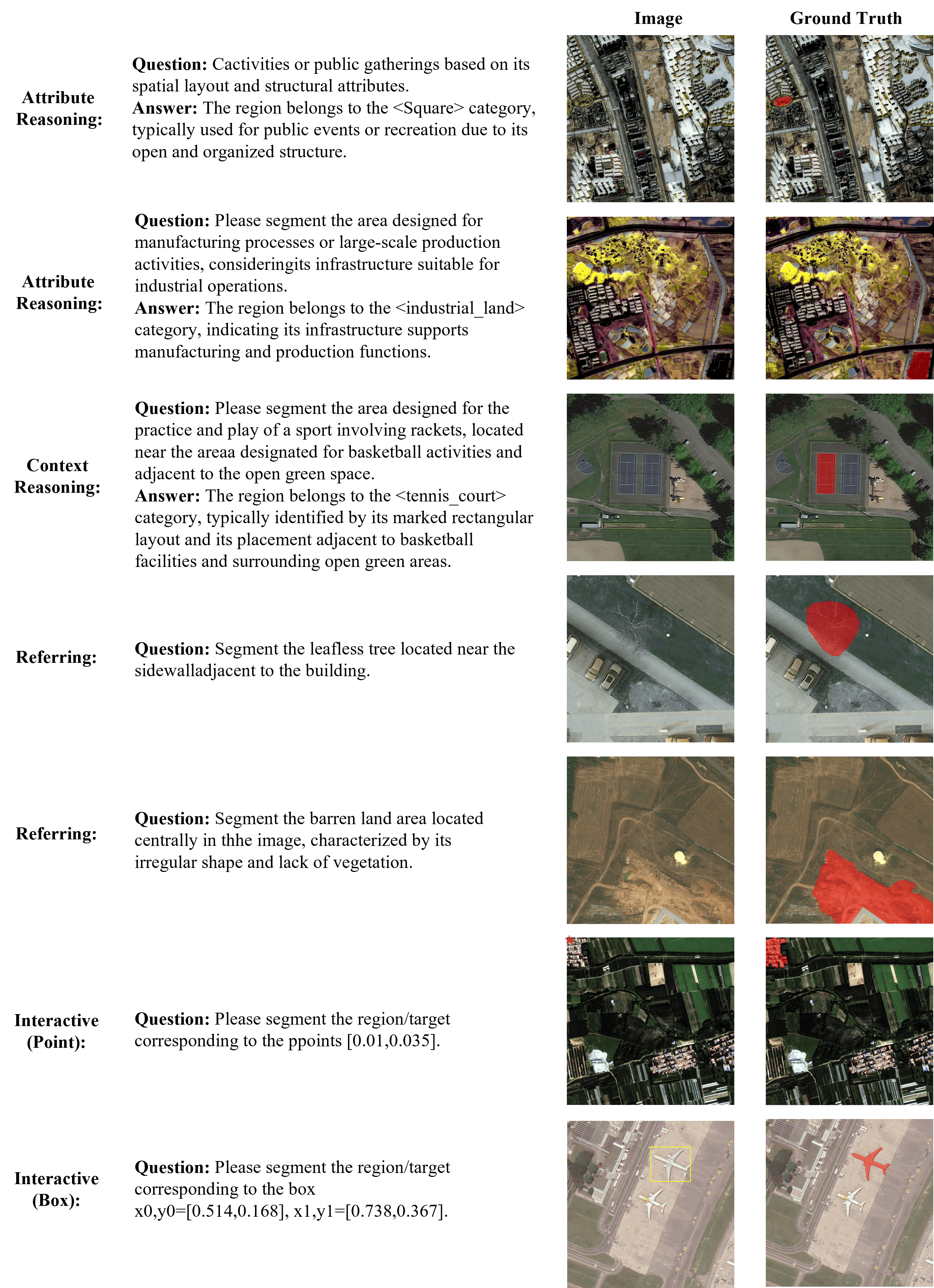}
    \caption{Additional samples of GeoSeg-1M.}
    \label{fig:supp_example_1m}
\end{figure*}

To complement the main paper, Fig.~\ref{fig:supp_example_1m} presents additional GeoSeg-1M samples that highlight the visual richness and instruction diversity of the dataset.

\section{More details about GeoSeg-Bench}
\label{sec:The statistics of GeoSeg-Bench}

GeoSeg-Bench comprises 6,892 samples organized into three subsets corresponding to reasoning, interactive, and referring segmentation tasks. The reasoning subset contains 1,711 samples, including 1,150 attribute reasoning instances and 561 contextual reasoning instances. The interactive subset includes 2,870 samples, with 1,435 samples using point-based prompts and 1,435 samples using box-based prompts. The referring subset consists of 2,311 samples. The benchmark covers 102 semantic categories. The full list of categories and the corresponding quantity distributions are provided in Tab.~\ref{tab:geoseg_bench_categories}. These statistics illustrate the overall diversity and balance of GeoSeg-Bench, supporting comprehensive evaluation across reasoning, referring, and interactive segmentation tasks.

\begin{figure}[ht]    
  \centering
  \includegraphics[width=\linewidth]{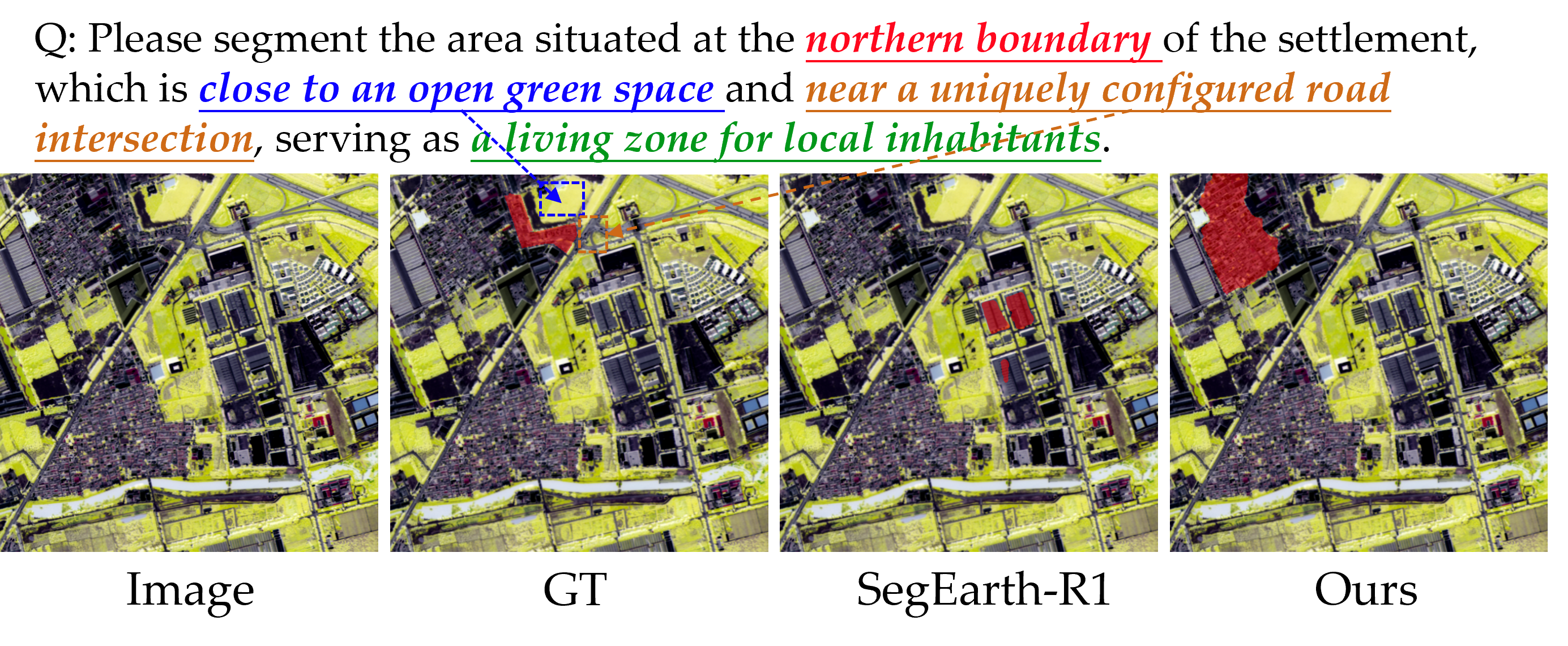}
  \caption{Example of a hard fail case.}
  \label{fig:rebuttal}
\end{figure}

Tab.~\ref{tab:rgram} reports instruction diversity on GeoSeg-1M and GeoSeg-Bench, excluding interactive template instructions. Referring instructions contain common words but maintain clear syntactic diversity, while reasoning instructions exhibit substantially higher linguistic diversity. The intra-class n-gram and syntactic diversity (Tab.~\ref{tab:rgram}, Row \textcolor{pink!200}{3}) indicates no obvious shortcut reasoning driven by superficial language patterns. In addition, Fig.~\ref{fig:rebuttal} shows a failure case in hard negative reasoning scene. Our model focuses on the \textcolor{red}{\underline{\textit{northern boundary}}} and \textcolor{green!60!black}{\underline{\textit{living zone}}} while ignoring the crucial cues \textcolor{blue}{\underline{\textit{close to an open green space}}} and \textcolor{orange!80!black}{\underline{\textit{near a road intersection}}}, which leads to a fail case.

\begin{table}[ht]
\caption{N-gram and Syntactic Diversity Statistics (\%).}
\label{tab:rgram}
\centering
\small
\resizebox{\columnwidth}{!}{
\begin{tabular}{lcccccc}
\hline
 & \multicolumn{2}{c}{2-gram} & \multicolumn{2}{c}{3-gram} & \multicolumn{2}{c}{Syntactic (POS)} \\
\cline{2-7}
 & Top50 & Top100 & Top50 & Top100 & Top50 & Top100 \\
\hline
GeoSeg-1M Referring & 53.89 & 64.56 & 37.89 & 47.85 & 36.83 & 43.23 \\
GeoSeg-1M Reasoning & 24.39 & 30.35 & 13.36 & 17.70 & 1.00 & 1.60 \\

\makecell[l]{GeoSeg-1M Class\\ Rural-Residential\\ (398 samples)} & 34.43 & 44.73 & 20.40 & 27.70 & 14.07 & 26.63 \\
GeoSeg-Bench Referring & 56.46 & 68.57 & 41.25 & 52.39 & 24.59 & 34.45 \\
GeoSeg-Bench Reasoning & 26.87 & 34.01 & 14.95 & 19.69 & 3.48 & 6.28 \\
\hline
\end{tabular}}
\end{table}

\begin{figure*}[th]
    \centering
    \includegraphics[width=0.88\linewidth]{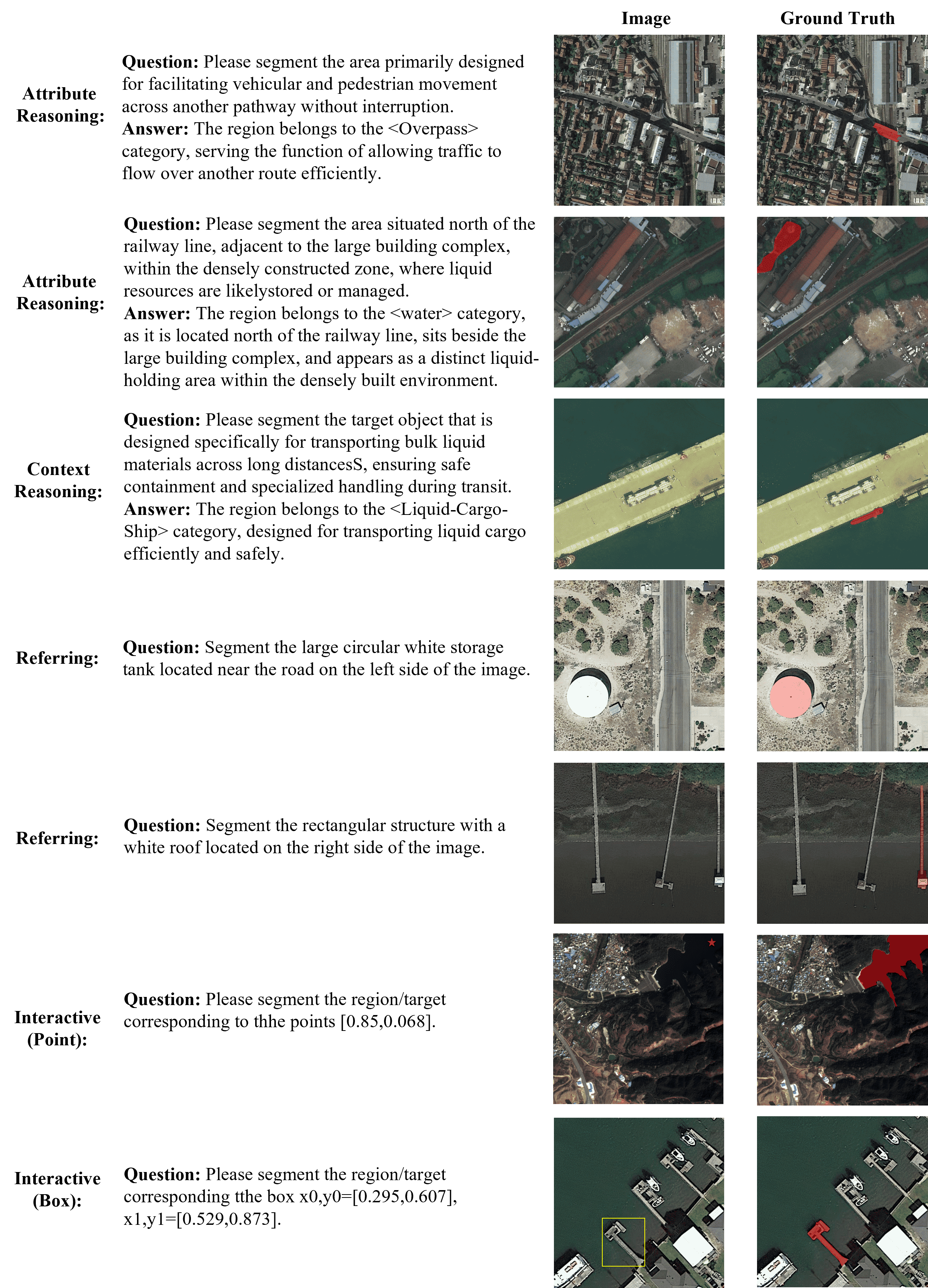}
    \caption{Additional samples of GeoSeg-Bench.}
    \label{fig:supp_example_bench}
\end{figure*}

To further illustrate the composition and annotation style of GeoSeg-Bench, we present additional representative samples in Fig.~\ref{fig:supp_example_bench}.

\section{More Details about Experiments}
\label{sec:More Details about Experiments}

\subsection{Implementation Details for UniGeoSeg}

\textbf{The subset with DIOR-derived data excluded.} To better examine the generalization behavior of UniGeoSeg under unfamiliar data sources, we additionally construct a subset of GeoSeg-1M by removing all samples originating from the DIOR~\cite{li2020object} family of datasets. The resulting subset contains 970,238 samples, comprising 109,274 reasoning, 431,204 referring, and 429,760 interactive instances. Since the downstream benchmarks SIOR~\cite{wang2023samrs} and DIOR-RSVG~\cite{zhan2023rsvg} are themselves derived from DIOR, excluding DIOR-based training data mitigates this source-specific overlap and provides a more independent setting for evaluating the model’s cross-task generalization.

\textbf{The subset used in ablations.} To conduct the ablation studies of task-adaptive text enhancement (TATE) and latent knowledge memory (LKM) under a manageable computational budget, we further construct a compact subset of 141,132 samples within the DIOR-excluded portion of GeoSeg-1M. This subset comprises 42,699 reasoning, 55,481 referring, and 42,952 interactive instances randomly sampled from the remaining data. We then evaluate the resulting models on the reasoning and referring tasks of GeoSeg-Bench, as well as on zero-shot interactive segmentation on SIOR, enabling a focused examination of how each module affects both the accuracy and the generalization capability of UniGeoSeg.

\textbf{Training Configuration.} All experiments are conducted with bfloat16 precision on eight NVIDIA A800 GPUs. Input images are resized to $512\times512$. We use the AdamW optimizer with an initial learning rate of $1\times10^{-4}$, a cosine decay schedule, and a warmup ratio of 0.03. Weight decay is set to 0.0. The progressive task scheduling (PTS) strategy is applied: the sampling weight of interactive segmentation gradually decreases to 0.7, with the remaining weight assigned to reasoning samples. Batch size is set to 16, and each experiment is trained for 3 epochs. Data loading uses 4 workers per GPU, and lazy preprocessing is enabled to accelerate training. Additional hyper-parameters and implementation details are summarized in Tab.~\ref{tab:train_hyperparams_supp}.

\begin{table*}[ht]
\centering
\small
\caption{Supplementary training hyper-parameters and setup.}
\label{tab:train_hyperparams_supp}
\begin{tabular}{ll|ll}
\toprule
\textbf{Parameter} & \textbf{Value} & \textbf{Parameter} & \textbf{Value} \\
\midrule
Precision & bfloat16 & GPUs & 8 × NVIDIA A800 \\
Image Size & $512\times512$ & Batch Size & 16 \\
Training Epochs & 3 & Optimizer & AdamW \\
Initial LR & $1\times10^{-4}$ & LR Schedule & Cosine decay \\
Warmup Ratio & 0.03 & Weight Decay & 0.0 \\
Gradient Checkpointing & Enabled & Data Loader Workers & 4 \\
PTS Sampling & Decays to 0.7 & ZeRO Stage & ZeRO 2 \\
\bottomrule
\end{tabular}
\end{table*}

\subsection{Comparison with Other Architectures}
We compare UniGeoSeg with several baseline models in terms of their visual encoder, language backbone, and the use of dual visual encoders. Notably, some baselines incorporate Segment Anything Model (SAM), employing a dual-encoder setup: one encoder provides image features to the segmentation decoder, while a second encoder aligned with the LLM backbone supplies features for language-guided understanding. UniGeoSeg, in contrast, does not use SAM and relies on a single visual encoder. Table~\ref{tab:model_architecture_comparison} summarizes these architectural differences across models.

\begin{table*}[htbp]
  \centering
  \caption{Comparison of UniGeoSeg with baseline model architectures. The ``Encoder Configuration'' column denotes the visual encoder setup: Dual-encoder indicates a SAM-based encoder for the segmentation decoder plus a second LLM-aligned encoder, while Single-encoder indicates only one visual encoder is used.}
  \label{tab:model_architecture_comparison}
  \begin{tabular}{@{}lccc@{}}
    \toprule
    Method & Vision Encoder & LLM Type & Encoder Configuration \\
    \midrule
    LISA\cite{lai2024lisa} & CLIP-L\cite{radford2021CLIP} & Vicuna-7B\cite{chiang2023vicuna} & Dual-encoder \\
    PixelLM\cite{ren2024pixellm} & CLIP-L & Vicuna-7B & Dual-encoder \\
    PSALM\cite{zhang2024psalm} & Swin-B\cite{liu2021swin} & Phi-1.5(1.3B)\cite{li2023phi15} & Single-encoder \\
    Geopixel\cite{shabbir2025geopixel} & CLIP-L & InternLM2-7B\cite{cai2024internlm2} & Dual-encoder \\
    Geopix\cite{ou2025geopix} & CLIP-L & Llama-2-7B\cite{touvron2023llama} & Single-encoder \\
    Earthmind\cite{shu2025earthmind} & InternViT\cite{chen2024internvl} & InternVL2-4B\cite{chen2024internvl} & Dual-encoder \\
    LISAt\cite{quenumlisat} & RemoteCLIP\cite{liu2024remoteclip} & Vicuna-7B & Dual-encoder \\
    Segearth-R1\cite{li2025segearthr1} & Swin-B & Phi-1.5(1.3B) & Single-encoder \\
    \cmidrule(lr){1-4}
    Ours (UniGeoSeg) & Swin-B & Phi-1.5(1.3B) & Single-encoder \\
    \bottomrule
  \end{tabular}
\end{table*}

Building on this comparison, UniGeoSeg employs a streamlined and efficient architecture. It relies on a single Swin-B\cite{liu2021swin} visual encoder,  reducing structural redundancy and computational overhead while maintaining strong visual feature representation. The language backbone of UniGeoSeg is relatively small, which further contributes to efficiency. Despite not using SAM encoder, UniGeoSeg achieves core SAM-like functionality by integrating interactive segmentation instructions and embedding visual prompts directly into the textual input space during training. Consequently, UniGeoSeg provides a compact and lightweight architecture that captures both language-guided and pixel-level cues, demonstrating advantages over both dual-encoder models and other single-encoder baselines.

\subsection{Fine-tuning Protocol} 
We summarize fine-tuning protocols on GeoSeg-1M for all baselines, as shown in Tab.~\ref{tab:finetune_protocol}.

\begin{table*}[!h]
\caption{Fine-tuning protocol for all baselines on GeoSeg-1M.}
\centering
\label{tab:finetune_protocol}
\resizebox{\textwidth}{!}{
\centering
\small
\setlength{\tabcolsep}{3pt}  
\begin{tabular}{lcccccccc}
\hline
Method & Epoch & Resolution & Encoder & Optimizer & Frozen & Finetune & LR & Batch \\
\hline
PSALM\cite{zhang2024psalm}        & 3 & 1024$\times$1024 & Single & AdamW  & \makecell[c]{vision encoder} & \makecell[c]{LLM, mask decoder} & 1e-4 & 8 \\
GeoPixel\cite{shabbir2025geopixel}     & 3 & 560$\times$560   & Dual   & AdamW  & \makecell[c]{vision encoder, SAM encoder, LLM} & \makecell[c]{vision/text projector mask decoder} & 3e-4 & 8 \\
Earthmind\cite{shu2025earthmind}    & 3 & 512$\times$512   & Dual   & Adam   & \makecell[c]{vision encoder, SAM encoder} & \makecell[c]{LLM-lora, projector, mask decoder} & 4e-5 & 16 \\
LISAT\cite{quenumlisat}        & 3 & 512$\times$512   & Dual   & AdamW  & \makecell[c]{vision encoder} & \makecell[c]{LLM-lora, projector, mask decoder} & 3e-4 & 16 \\
SegEarth-R1\cite{li2025segearthr1}  & 3 & 1024$\times$1024 & Single & AdamW  & \makecell[c]{vision encoder} & \makecell[c]{LLM, mask decoder} & 1e-4 & 8 \\
UniGeoSeg    & 3 & 512$\times$512   & Single & AdamW  & \makecell[c]{vision encoder} & \makecell[c]{LLM, mask decoder} & 1e-4 & 8 \\
\hline
\end{tabular}}
\end{table*}

\section{Additional Examples of Model Predictions}
\label{sec:Additional Examples of Model Predictions}

\begin{figure*}[th]
    \centering
    \includegraphics[width=\linewidth]{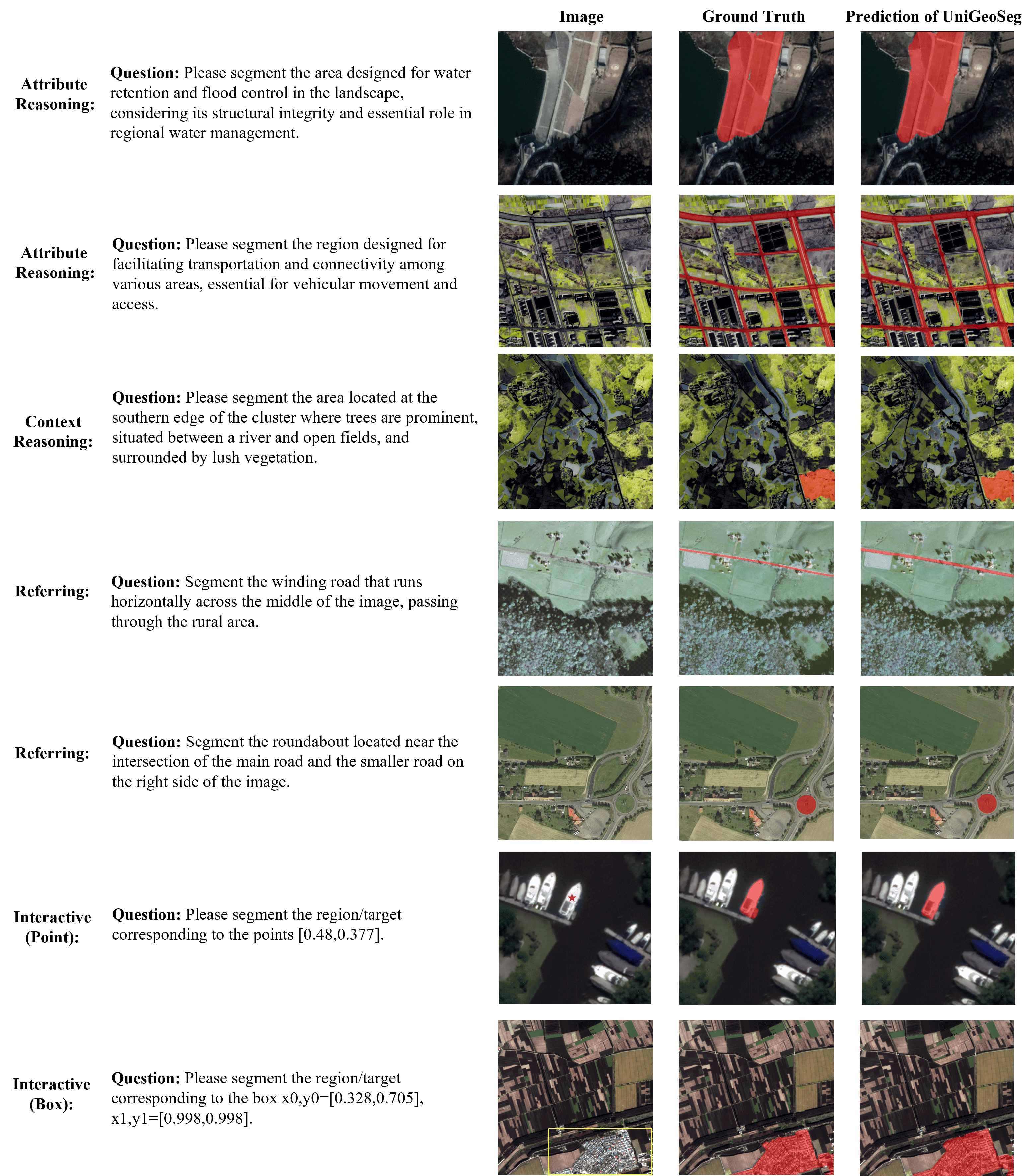}
    \caption{Additional samples of model prediction by UniGeoSeg.}
    \label{fig:supp_prediction}
\end{figure*}

This section presents additional qualitative results to further illustrate the behavior of the model in various scenarios, as shown in Fig.~\ref{fig:supp_prediction}.

\section{Evaluating UniGeoSeg with Alternative Language Model}
\label{sec:Evaluating UniGeoSeg with Alternative Language Model}

In this section, we examine the flexibility of UniGeoSeg by replacing its original Phi-1.5~\cite{li2023phi15} language backbone with DeepSeek-7B-Chat~\cite{bi2024deepseek}. The model is trained on GeoSeg-1M under the same settings as the main configuration and evaluated on the EarthReason~\cite{li2025segearthr1} and RRSIS-D~\cite{liu2024rotated} benchmarks. While switching the language backbone naturally introduces some variation in overall performance, the DeepSeek-based variant still maintains competitive segmentation quality.

To further validate the general applicability of our proposed modules TATE, LKM, and PTS, we perform additional ablation studies using an alternative backbone. Following identical training and evaluation protocols, we remove each module in turn and report the results in Tab.~\ref{tab:deepseek}. The consistent performance gains observed for UniGeoSeg across different language model architectures indicate that these modules are largely model-agnostic and provide robust improvements independent of the underlying LLM.

\begin{table*}[ht]
  \caption{Results on EarthReason and RRSIS-D of UniGeoSeg-DeepSeek7B.}
  \label{tab:deepseek}
  \centering
  \resizebox{\textwidth}{!}{
  \begin{tabular}{@{}lcccccc@{}}
    \toprule
    \multirow{2}{*}{Method} 
      & \multicolumn{2}{c}{EarthReason (Val)} 
      & \multicolumn{2}{c}{EarthReason (Test)} 
      & \multicolumn{2}{c}{RRSIS-D} \\
    \cmidrule(lr){2-3} \cmidrule(lr){4-5} \cmidrule(lr){6-7}
      & cIoU & gIoU & cIoU & gIoU & cIoU & gIoU \\
    \midrule

    \makecell[l]{UniGeoSeg-DeepSeek7B \\(w/o TATE, LKM, and PTS)} & 67.37 & 65.03 & 65.92 & 64.23 & 71.52 & 60.02 \\

    \midrule
    UniGeoSeg-DeepSeek7B & 71.41 \textbf{(+4.04)} & 68.88 \textbf{(+3.85)} & 71.93 \textbf{(+6.01)} & 69.17 \textbf{(+4.94)} & 75.12 \textbf{(+3.60)} & 65.75 \textbf{(+5.73)} \\
    \bottomrule
  \end{tabular}
  }
\end{table*}

\section{Datasheets}
\label{app-datasheets}

\subsection{Motivation}

\begin{enumerate}
    \item \textit{``For what purpose was the dataset created?''}
    
    \textcolor{BurntOrange}{\textbf{A:}} 
The GeoSeg-1M and GeoSeg-Bench datasets were created to address the lack of large-scale, high-quality multimodal segmentation data in remote sensing, a bottleneck that limits the development of unified vision–language segmentation models. GeoSeg-1M provides a million-level collection of image–mask–instruction triplets covering referring, reasoning, and interactive segmentation, enabling instruction-driven multimodal understanding at scale. GeoSeg-Bench complements it with a carefully designed evaluation suite for assessing fine-grained reasoning ability, context-aware grounding, and interactive segmentation performance. Together, these two datasets aim to support the training and rigorous benchmarking of general-purpose remote sensing segmentation models capable of robust instruction following and comprehensive scene understanding.
    
    \item \textit{``Who created the dataset (\textit{e.g.}, which team, research group) and on behalf of which entity?''}
    
    \textcolor{BurntOrange}{\textbf{A:}} The dataset was created by the following authors:
    \begin{itemize}
      \item Anonymous authors
    \end{itemize}
    
    \item \textit{``Who funded the creation of the dataset?''}
    
    \textcolor{BurntOrange}{\textbf{A:}}
    The dataset creation was funded by the affiliations of the authors involved in this work.
\end{enumerate}

\subsection{Composition}
Most of the questions in this section are intended to provide dataset consumers with the information they need to make informed decisions about using the dataset for their chosen tasks. Some of the questions are designed to elicit information about compliance with the EU’s General Data Protection Regulation (GDPR) or comparable regulations in other jurisdictions. Questions that apply only to datasets that relate to people are grouped together at the end of the section. We recommend taking a broad interpretation of whether a dataset relates to people. For example, any dataset containing text that was written by people relates to people.
\begin{enumerate}
    \item \textit{``What do the instances that comprise our datasets represent (\textit{e.g.}, documents, photos, people, countries)?''}
    
    \textcolor{BurntOrange}{\textbf{A:}} The dataset primarily consists of remote sensing images captured by satellites and drones, along with their corresponding textual annotations. All datasets utilized in GeoSeg-1M and GeoSeg-Bench are publicly accessible and nonprofit.
    
    \item \textit{``How many instances are there in total (of each type, if appropriate)?''}
    
    \textcolor{BurntOrange}{\textbf{A:}} GeoSeg-1M includes 1,148,504 image-mask-instruction triplets . Details could be found in the main text. GeoSeg-Bench consists of 6,892 samples, including 2,870 interactive, 2,311 referring, and 1,711 reasoning segmentation samples.

    \item \textit{``Does the dataset contain all possible instances or is it a sample (not necessarily random) of instances from a larger set?''}
    
    \textcolor{BurntOrange}{\textbf{A:}} The images in GeoSeg-1M and GeoSeg-Bench are sourced from existing detection and segmentation  datasets. Except for the samples from RemoteSAM~\cite{yao2025remotesam}, RRSIS-D~\cite{liu2024rotated}, and EarthReason~\cite{li2025segearthr1} in GeoSeg-1M, all textual annotations were independently created by us.
    
    \item \textit{``Is there a label or target associated with each instance?''}
    
    \textcolor{BurntOrange}{\textbf{A:}} Yes, for these images, we have provided image-mask-instruction triplets instances.
    
    \item \textit{``Is any information missing from individual instances?''}
    
    \textcolor{BurntOrange}{\textbf{A:}} No, each individual instance is complete.
    
    \item \textit{``Are relationships between individual instances made explicit (\textit{e.g.}, users’ movie ratings, social network links)?''}
    
    \textcolor{BurntOrange}{\textbf{A:}} Yes, the relationship between individual instances is explicit.
    
    \item \textit{``Are there recommended data splits (\textit{e.g.}, training, development/validation, testing)?''}
    
    \textcolor{BurntOrange}{\textbf{A:}} 
    The GeoSeg-1M is designed to train the RS MLLMs for instruction-driven segmentation, and the GeoSeg-Bench is designed to evaluation.
    
    \item \textit{``Is the dataset self-contained, or does it link to or otherwise rely on external resources (\textit{e.g.}, websites, tweets, other datasets)?''}
    
    \textcolor{BurntOrange}{\textbf{A:}} GeoSeg-1M and GeoSeg-Bench are self-contained and will be open-sourced on platforms like Hugging Face for easy use.
    
    \item \textit{``Does the dataset contain data that might be considered confidential (\textit{e.g.}, data that is protected by legal privilege or by doctor–patient confidentiality, data that includes the content of individuals’ non-public communications)?''}
    
    \textcolor{BurntOrange}{\textbf{A:}} No, all data are clearly licensed.
    
    \item \textit{``Does the dataset contain data that, if viewed directly, might be offensive, insulting, threatening, or might otherwise cause anxiety?''}
    
    \textcolor{BurntOrange}{\textbf{A:}} No, GeoSeg-1M and GeoSeg-Bench do not contain any data with negative information.
\end{enumerate}

\subsection{Collection Process}
In addition to the goals outlined in the previous section, the questions in this section are designed to elicit information that may help researchers and practitioners create alternative datasets with similar characteristics. Again, questions that apply only to datasets that relate to people are grouped together at the end of the section.
\begin{enumerate}
    \item \textit{``How was the data associated with each instance acquired?''}
    
    \textcolor{BurntOrange}{\textbf{A:}} 
    The images in GeoSeg-1M and GeoSeg-Bench are sourced from existing detection  and segmentation datasets. We enrich these with annotations. Details are shown in the Section 3 in main text.
    
    \item \textit{``What mechanisms or procedures were used to collect the data (\textit{e.g.}, hardware apparatuses or sensors, manual human curation, software programs, software APIs)?''}
    
    \textcolor{BurntOrange}{\textbf{A:}} GeoSeg-1M is constructed by integrating multiple publicly available remote sensing datasets collected from airborne and satellite imaging platforms with diverse spatial resolutions. All images and pixel-level masks originate from their respective sources without additional sensor deployment. Data consolidation, preprocessing, and annotation standardization were performed through automated software pipelines, including unified format conversion, image tiling, connected-region decomposition, and large-scale mask filtering. Instruction annotations were generated using GPT-4o through API-based prompting, followed by cross-evaluation with open-source vision-language models. No manual labeling was introduced beyond quality verification of a small subset for prompt and pipeline validation. GeoSeg-Bench was further curated by two domain experts in remote sensing, who independently reviewed and cross-validated candidate samples to ensure high-quality and unambiguous ground truth.
    
    \item \textit{``If the dataset is a sample from a larger set, what was the sampling strategy (\textit{e.g.}, deterministic, probabilistic with specific sampling probabilities)?''} 
    
    \textcolor{BurntOrange}{\textbf{A:}} Please refer to the details listed in the main text Section 3.
\end{enumerate}

\subsection{Preprocessing, Cleaning, and Labeling}
The questions in this section are intended to provide dataset
consumers with the information they need to determine whether the “raw” data has been processed in ways that are compatible with their chosen tasks. For example, text that has been converted into a ``bag-of-words" is not suitable for tasks involving word order.
\begin{enumerate}
    \item \textit{``Was any preprocessing/cleaning/labeling of the data done (\textit{e.g.}, discretization or bucketing, tokenization, part-of-speech tagging, SIFT feature extraction, removal of instances, processing of missing values)?''}
    
    \textcolor{BurntOrange}{\textbf{A:}} Yes. Extensive preprocessing and cleaning were applied to unify heterogeneous remote sensing datasets. All images were tiled into 512$\times$512 patches with a stride of 256, and masks were converted into a unified binary format. Each mask was decomposed into connected regions to isolate individual objects. A two-stage filtering pipeline was used to remove low-quality or noisy regions, combining rule-based screening with model-based quality assessment. Categories across datasets were merged into a harmonized taxonomy. Instruction annotations were automatically generated and subsequently cross-evaluated by large vision–language models. No manual relabeling was added beyond limited spot-checking for validation.

    \item \textit{``Was the `raw' data saved in addition to the preprocessed/cleaned/labeled data (\textit{e.g.}, to support unanticipated future uses)?''} 
    
    \textcolor{BurntOrange}{\textbf{A:}} Yes, raw data is accessible.
    
    \item \textit{``Is the software that was used to preprocess/clean/label the data available?''} 
    
    \textcolor{BurntOrange}{\textbf{A:}} Yes, the necessary software used to preprocess and clean the data is publicly available.
\end{enumerate}

\subsection{Uses}
The questions in this section are intended to encourage dataset creators to reflect on tasks for which the dataset should and should not be used. By explicitly highlighting these tasks, dataset creators can help dataset consumers make informed decisions, thereby avoiding potential risks or harms.
\begin{enumerate}
    \item \textit{``Has the dataset been used for any tasks already?''} 
    
    \textcolor{BurntOrange}{\textbf{A:}} 
    No.
    
    \item \textit{``Is there a repository that links to any or all papers or systems that use the dataset?''} 
    
    \textcolor{BurntOrange}{\textbf{A:}} Yes, we will provide such links in the GitHub and the Huggingface repository.
    
    \item \textit{``What (other) tasks could the dataset be used for?''} 
    
    \textcolor{BurntOrange}{\textbf{A:}} GeoSeg-1M provides extensive annotations for instruction-driven segmentation tasks. It could be used to train the MLLMs. GeoSeg-Bench provides high-quality samples in interactive, referring, and reasoning segmentation tasks. It could be used to evaluate the MLLMs.
    
    \item \textit{``Is there anything about the composition of the dataset or the way it was collected and preprocessed/cleaned/labeled that might impact future uses?''} 
    
    \textcolor{BurntOrange}{\textbf{A:}} No.
    
    \item \textit{``Are there tasks for which the dataset should not be used?''} 
    
    \textcolor{BurntOrange}{\textbf{A:}} N/A.
\end{enumerate}

\subsection{Distribution}
Dataset creators should provide answers to these questions prior to distributing the dataset either internally within the entity on behalf of which the dataset was created or externally to third parties.
\begin{enumerate}
    \item \textit{``Will the dataset be distributed to third parties outside of the entity (\textit{e.g.}, company, institution, organization) on behalf of which the dataset was created?''} 
    
    \textcolor{BurntOrange}{\textbf{A:}} The datasets will be made publicly accessible to the research community.
    
    \item \textit{``How will the dataset be distributed (\textit{e.g.}, tarball on website, API, GitHub)?''} 
    
    \textcolor{BurntOrange}{\textbf{A:}} We will provide GeoSeg-1M and GeoSeg-Bench in the GitHub and the Huggingface repository.
    
    \item \textit{``When will the dataset be distributed?''} 
    
    \textcolor{BurntOrange}{\textbf{A:}} We will create a repository to release the data once the paper is officially published.
    
    \item \textit{``Will the dataset be distributed under a copyright or other intellectual property (IP) license, and/or under applicable terms of use (ToU)?''} 
    
    \textcolor{BurntOrange}{\textbf{A:}} Yes, the dataset will be released under the Creative Commons Attribution-NonCommercial-ShareAlike 4.0 International License.
    
    \item \textit{``Have any third parties imposed IP-based or other restrictions on the data associated with the instances?''} 
    
    \textcolor{BurntOrange}{\textbf{A:}} No.
    
    \item \textit{``Do any export controls or other regulatory restrictions apply to the dataset or to individual instances?''} 
    
    \textcolor{BurntOrange}{\textbf{A:}} No.    
\end{enumerate}

\subsection{Maintenance}
As with the questions in the previous section, dataset creators should provide answers to these questions prior to distributing the dataset. The questions in this section are intended to encourage dataset creators to plan for dataset maintenance and communicate this plan to dataset consumers.
\begin{enumerate}
    \item \textit{``Who will be supporting/hosting/maintaining the dataset?''} 
    
    \textcolor{BurntOrange}{\textbf{A:}} The authors of this work serve to support, host, and maintain the datasets.
    
    \item \textit{``How can the owner/curator/manager of the dataset be contacted (\textit{e.g.}, email address)?''} 
    
    \textcolor{BurntOrange}{\textbf{A:}} They can be contacted via the email addresses listed on the paper or webpage.
    
    \item \textit{``Is there an erratum?''} 
    
    \textcolor{BurntOrange}{\textbf{A:}} There is no explicit erratum; updates and known errors will be specified in future versions.
    
    \item \textit{``Will the dataset be updated (\textit{e.g.}, to correct labeling errors, add new instances, delete instances)?''} 
    
    \textcolor{BurntOrange}{\textbf{A:}} Future updates (if any) will be posted on the dataset website.
    
    \item \textit{``Will older versions of the dataset continue to be supported/hosted/maintained?''} 
    
    \textcolor{BurntOrange}{\textbf{A:}} 

    Yes. This initial release will be updated in the future, with older versions replaced as new updates are posted.
    
    \item \textit{``If others want to extend/augment/build on/contribute to the dataset, is there a mechanism for them to do so?''} 
    
    \textcolor{BurntOrange}{\textbf{A:}} Yes, we will provide detailed instructions for future extensions.
\end{enumerate}

\begin{table*}[ht]
\centering
\small
\caption{The 117 semantic categories included in GeoSeg-1M.}
\label{tab:geoseg_categories}
\resizebox{\textwidth}{!}{%
\begin{tabular}{lll}
\toprule
1. Urban\_Land & 2. Agriculture\_Land & 3. Rangeland \\
4. Forest\_Land & 5. Water & 6. Barren\_Land \\
7. Low Vegetation / Field & 8. Impervious (Other) & 9. Road \\
10. Industrial\_Land & 11. Urban\_Residential & 12. Rural\_Residential \\
13. Paddy\_Field & 14. Irrigated\_Land & 15. Dry\_Cropland \\
16. Garden\_Plot & 17. Arbor\_Woodland & 18. Shrub\_Land \\
19. Natural\_Grassland & 20. Artificial\_Grassland & 21. River \\
22. Lake & 23. Pond & 24. Background \\
25. Building & \makecell[l]{26. Industrial/Commercial/Public/\\Military/Private/Transport Units} & 27. Mine/Dump/Construction Sites \\
28. Artificial Non-Agricultural Vegetated Areas & 29. Arable Land (Annual Crops) & 30. Permanent Crops \\
31. Pastures & 32. Complex/Mixed Cultivation Patterns & 33. Orchards at Urban Fringe \\
34. Herbaceous Vegetation Associations & 35. Open Spaces w/ Little Vegetation & 36. Wetlands \\
37. Clouds/Shadows & 38. Park & 39. Snow \\
40. Fish Pond & 41. Stadium & 42. Square \\
43. Overpass & 44. Railway Station & 45. Airport \\
46. Pervious Surface & 47. Coniferous & 48. Deciduous \\
49. Brushwood & 50. Vineyard & 51. Plowed Land \\
52. Swimming Pool & 53. Clear Cut & 54. Mixed \\
55. Ligneous & 56. Greenhouse & 57. Cropland \\
58. Grass & 59. Tundra & 60. Impervious \\
61. Low Vegetation & 62. Tree & 63. Car \\
64. Large-Vehicle & 65. Swimming-Pool & 66. Helicopter \\
67. Bridge & 68. Plane & 69. Ship \\
70. Soccer-Ball-Field & 71. Basketball-Court & 72. Ground-Track-Field \\
73. Small-Vehicle & 74. Baseball-Diamond & 75. Tennis-Court \\
76. Roundabout & 77. Storage-Tank & 78. Harbor \\
79. Container-Crane & 80. Airport & 81. Helipad \\
82. Chimney & 83. Expressway-Service-Area & 84. Expressway-Toll-Station \\
85. Dam & 86. GolfField & 87. Windmill \\
88. A220 & 89. A321 & 90. A330 \\
91. A350 & 92. ARJ21 & 93. Boeing737 \\
94. Boeing747 & 95. Boeing777 & 96. Boeing787 \\
97. Bus & 98. C919 & 99. Cargo-Truck \\
100. Dry-Cargo-Ship & 101. Dump-Truck & 102. Engineering-Ship \\
103. Excavator & 104. Fishing-Boat & 105. Intersection \\
106. Liquid-Cargo-Ship & 107. Motorboat & 108. Other-Airplane \\
109. Other-Ship & 110. Other-Vehicle & 111. Passenger-Ship \\
112. Tractor & 113. Trailer & 114. Truck-Tractor \\
115. Tugboat & 116. Van & 117. Warship \\
\bottomrule
\end{tabular}
}
\end{table*}

\begin{table*}[htbp]
\centering
\caption{Category Statistics of GeoSeg-Bench}
\label{tab:geoseg_bench_categories}
\resizebox{\textwidth}{!}{%
\begin{tabular}{lll|lll|lll}
\hline
\textbf{ID} & \textbf{Name} & \textbf{Count (\%)} & 
\textbf{ID} & \textbf{Name} & \textbf{Count (\%)} & 
\textbf{ID} & \textbf{Name} & \textbf{Count (\%}) \\
\hline
1 & Urban\_Land & 37 (0.53\%) & 2 & Agriculture\_Land & 207 (2.97\%) & 3 & Rangeland & 24 (0.34\%) \\
4 & Forest\_Land & 97 (1.39\%) & 5 & Water & 260 (3.74\%) & 6 & Barren\_Land & 226 (3.25\%) \\
7 & Low Vegetation / Field & 51 (0.73\%) & 8 & Impervious (Other) & 2 (0.03\%) & 9 & Road & 478 (6.87\%) \\
10 & Industrial\_Land & 329 (4.73\%) & 11 & Urban\_Residential & 196 (2.82\%) & 12 & Rural\_Residential & 378 (5.43\%) \\
13 & Paddy\_Field & 118 (1.70\%) & 14 & Irrigated\_Land & 398 (5.72\%) & 15 & Dry\_Cropland & 77 (1.11\%) \\
16 & Garden\_Plot & 101 (1.45\%) & 17 & Arbor\_Woodland & 176 (2.53\%) & 18 & Shrub\_Land & 69 (0.99\%) \\
19 & Natural\_Grassland & 92 (1.32\%) & 20 & Artificial\_Grassland & 82 (1.18\%) & 21 & River & 240 (3.45\%) \\
22 & Lake & 85 (1.22\%) & 23 & Pond & 149 (2.14\%) & 24 & Background & 11 (0.16\%) \\
25 & Building & 154 (2.21\%) & 26 & \makecell[l]{Industrial, Commercial, \\Public, Military, Private\\ And Transport Units} & 51 (0.73\%) & 27 & \makecell[l]{Mine, Dump And\\ Construction Sites} & 17 (0.24\%) \\
28 & \makecell[l]{Artificial Non-Agricultural \\Vegetated Areas} & 30 (0.43\%) & 29 & \makecell[l]{Arable Land \\(Annual Crops)} & 29 (0.42\%) & 30 & Permanent Crops & 18 (0.26\%) \\
31 & Pastures & 50 (0.72\%) & 34 & \makecell[l]{Herbaceous Vegetation\\ Associations} & 31 (0.45\%) & 35 & \makecell[l]{Open Spaces With Little\\ Or No Vegetation} & 24 (0.34\%) \\
36 & Wetlands & 31 (0.45\%) & 38 & Park & 10 (0.14\%) & 39 & Snow & 15 (0.22\%) \\
40 & Fish Pond & 13 (0.19\%) & 41 & Stadium & 34 (0.49\%) & 42 & Square & 13 (0.19\%) \\
43 & Overpass & 76 (1.09\%) & 44 & Railway Station & 27 (0.39\%) & 45 & Airport & 3 (0.04\%) \\
46 & Pervious Surface & 29 (0.42\%) & 47 & Coniferous & 10 (0.14\%) & 48 & Deciduous & 29 (0.42\%) \\
49 & Brushwood & 19 (0.27\%) & 50 & Vineyard & 20 (0.29\%) & 51 & Plowed Land & 19 (0.27\%) \\
52 & Swimming Pool & 37 (0.53\%) & 53 & Clear Cut & 3 (0.04\%) & 56 & Greenhouse & 8 (0.11\%) \\
57 & Cropland & 92 (1.32\%) & 58 & Grass & 67 (0.96\%) & 60 & Impervious & 59 (0.85\%) \\
61 & Low Vegetation & 15 (0.22\%) & 62 & Tree & 50 (0.72\%) & 63 & Car & 62 (0.89\%) \\
64 & Large-Vehicle & 134 (1.93\%) & 65 & Swimming-Pool & 37 (0.53\%) & 66 & Helicopter & 5 (0.07\%) \\
67 & Bridge & 67 (0.96\%) & 68 & Plane & 304 (4.37\%) & 69 & Ship & 151 (2.17\%) \\
70 & Soccer-Ball-Field & 65 (0.93\%) & 71 & Basketball-Court & 72 (1.03\%) & 72 & Ground-Track-Field & 47 (0.68\%) \\
73 & Small-Vehicle & 65 (0.93\%) & 74 & Baseball-Diamond & 85 (1.22\%) & 75 & Tennis-Court & 245 (3.52\%) \\
76 & Roundabout & 53 (0.76\%) & 77 & Storage-Tank & 80 (1.15\%) & 78 & Harbor & 199 (2.86\%) \\
79 & Container-Crane & 6 (0.09\%) & 80 & Airport & 45 (0.65\%) & 82 & Chimney & 20 (0.29\%) \\
83 & Expressway-Service-Area & 8 (0.11\%) & 84 & Expressway-Toll-Station & 55 (0.79\%) & 85 & Dam & 19 (0.27\%) \\
86 & GolfField & 10 (0.14\%) & 87 & Windmill & 3 (0.04\%) & 90 & A330 & 4 (0.06\%) \\
94 & Boeing747 & 3 (0.04\%) & 95 & Boeing777 & 1 (0.01\%) & 96 & Boeing787 & 1 (0.01\%) \\
97 & Bus & 3 (0.04\%) & 99 & Cargo-Truck & 15 (0.22\%) & 100 & Dry-Cargo-Ship & 32 (0.46\%) \\
101 & Dump-Truck & 14 (0.20\%) & 102 & Engineering-Ship & 4 (0.06\%) & 103 & Excavator & 4 (0.06\%) \\
104 & Fishing-Boat & 2 (0.03\%) & 105 & Intersection & 22 (0.32\%) & 106 & Liquid-Cargo-Ship & 13 (0.19\%) \\
107 & Motorboat & 7 (0.10\%) & 108 & Other-Airplane & 1 (0.01\%) & 110 & Other-Vehicle & 1 (0.01\%) \\
111 & Passenger-Ship & 10 (0.14\%) & 113 & Trailer & 3 (0.04\%) & 114 & Truck-Tractor & 3 (0.04\%) \\
115 & Tugboat & 1 (0.01\%) & 116 & Van & 5 (0.07\%) & 117 & Warship & 6 (0.09\%) \\
\hline
\end{tabular}}
\end{table*}

\end{document}